\documentclass[10pt,twocolumn,letterpaper]{article}

\usepackage{wacv}              

\usepackage[dvipsnames]{xcolor}
\usepackage[switch]{lineno}
\usepackage{algorithmic}
\usepackage{algorithm}
\usepackage{amsmath}
\usepackage{amssymb}
\usepackage{amsthm}
\usepackage{booktabs}
\usepackage{graphicx}
\usepackage{multirow}
\usepackage{xspace}

\usepackage[pagebackref,breaklinks,colorlinks]{hyperref}

\newcommand{\red}[1]{{\color{red}#1}}
\newcommand{\refred}[1]{{\color{red}#1}}

\newcommand{\bfgreen}[1]{{\textbf{{\color[RGB]{0,153,51}#1}}}}
\newcommand{\bfred}[1]{{\textbf{{\color[RGB]{255,100,97}#1}}}}
\newcommand{\methodname}{\emph{DyRoNet}\xspace}

\usepackage[capitalize]{cleveref}
\crefname{section}{Sec.}{Secs.}
\Crefname{section}{Section}{Sections}
\Crefname{table}{Table}{Tables}
\crefname{table}{Tab.}{Tabs.}

\def\wacvPaperID{2029} 
\def\confName{WACV}
\def\confYear{2025}

\begin{document}
\title{DyRoNet: Dynamic Routing and Low-Rank Adapters for Autonomous Driving Streaming Perception}

\author{
Xiang Huang\textsuperscript{1}\thanks{This work was completed during a visit to CMU and Alibaba.}, 
Zhi-Qi Cheng\textsuperscript{2}\thanks{Corresponding author, also a Visiting Assistant Professor at CMU.}, 
Jun-Yan He\textsuperscript{3}, 
Chenyang Li\textsuperscript{3}, 
Wangmeng Xiang\textsuperscript{3,4}, 
Baigui Sun\textsuperscript{3} \vspace{0.3em} \\
\textsuperscript{1}Institute of Artificial Intelligence, Southwest Jiaotong University, Chengdu, China \\
\textsuperscript{2}School of Engineering and Technology, University of Washington, Tacoma, WA, USA \\
\textsuperscript{3}Institute for Intelligent Computing, Alibaba Group, Shenzhen, China \\
\textsuperscript{4}Department of Computing, The Hong Kong Polytechnic University, Hong Kong \vspace{0.3em}\\
{\tt\small xianghuang@my.swjtu.edu.cn, zhiqics@uw.edu, junyanhe1989@gmail.com,} \\
{\tt\small lichenyang.scut@foxmail.com, marquezxm@gmail.com, sunbaigui85@126.com}
\vspace{-1.5em}
}

\maketitle
%
\begin{abstract}
The advancement of autonomous driving systems hinges on the ability to achieve low-latency and high-accuracy perception.~To address this critical need, this paper introduces \textbf{Dy}namic \textbf{Ro}utering \textbf{Net}work (\methodname), a low-rank enhanced dynamic routing framework designed for streaming perception in autonomous driving systems.~\methodname integrates a suite of pre-trained branch networks, each meticulously fine-tuned to function under distinct environmental conditions.~At its core, the framework offers a speed router module, developed to assess and route input data to the most suitable branch for processing.~This approach not only addresses the inherent limitations of conventional models in adapting to diverse driving conditions but also ensures the balance between performance and efficiency.~Extensive experimental evaluations demonstrating the adaptability of \methodname to diverse branch selection strategies, resulting in significant performance enhancements across different scenarios.~This work not only establishes a new benchmark for streaming perception but also provides valuable engineering insights for future work.\footnote{Project:~\url{https://tastevision.github.io/DyRoNet/}}
\end{abstract}
 
\vspace{-2mm}
\section{Introduction}
\label{sec:intro}
In autonomous driving systems, it is crucial to achieve low-latency and high-precision perception. Traditional object detection algorithms~\cite{zou2023object}, while effective in various contexts, often confront the challenge of latency due to inherent computational delays. This lag between algorithmic processing and real-world states can lead to notable discrepancies between predicted and actual object locations. Such latency issues have been extensively reported and are known to significantly impact the decision-making process in autonomous driving systems~\cite{chen2023endtoend}.

Addressing these challenges, the concept of streaming perception has been introduced as a response~\cite{li2020towards}. This perception task aims to predict ``future" results by accounting for the delays incurred during the frame processing stage. Unlike traditional methods that primarily focus on detection at a given moment, streaming perception transcends this limitation by anticipating future environmental states, and aligning perceptual outputs closer to real-time dynamics. This new paradigm is key in addressing the critical gap between real-time processing and real-world changes, thereby enhancing the safety and reliability of autonomous driving systems~\cite{muhammad2020deep}.

\begin{figure}[!t]
    \centering
    \includegraphics[width=0.95\linewidth]{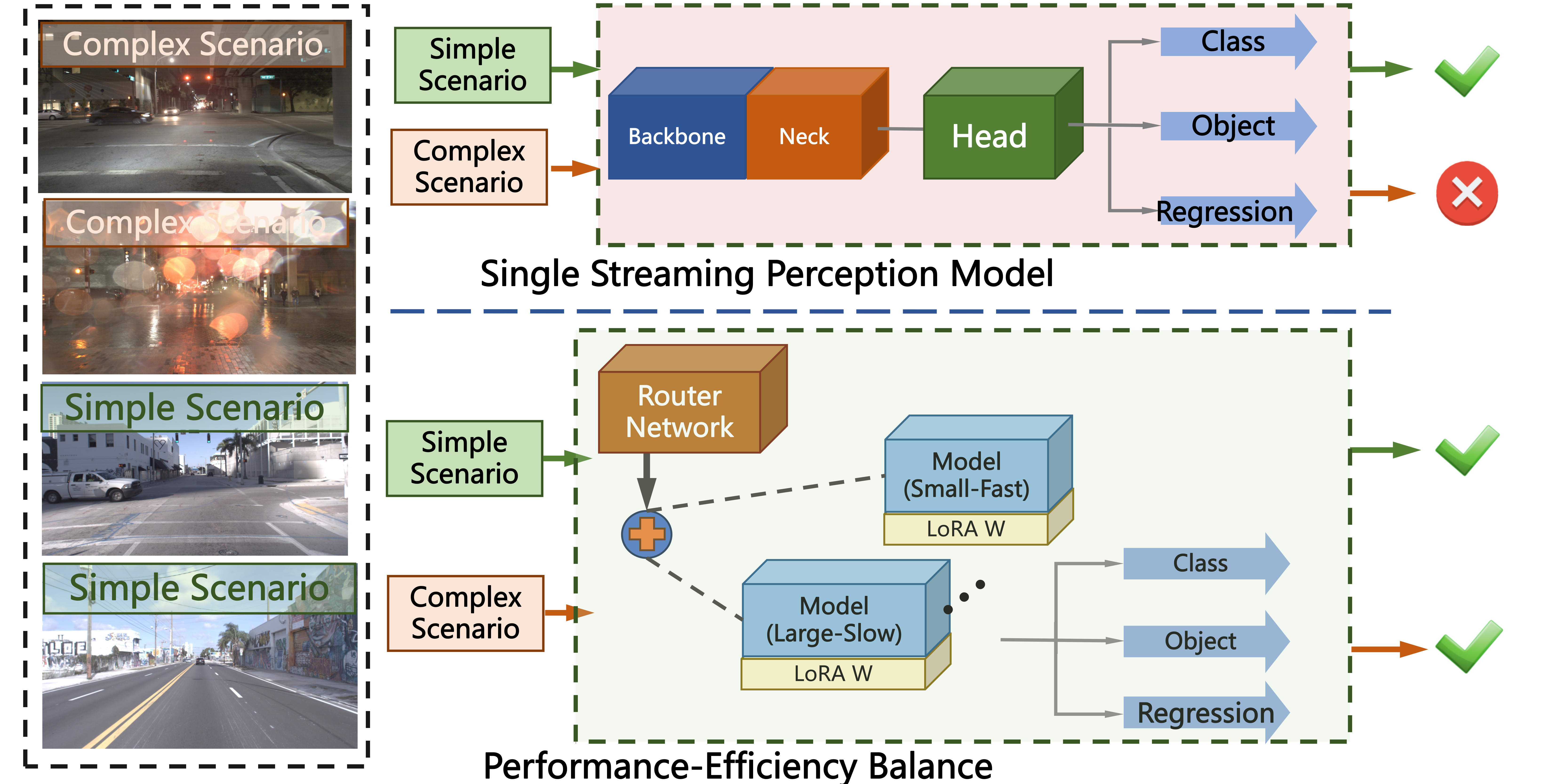}
    \caption{\small Illustration of DyRoNet's adaptive selection mechanism in streaming perception, contrasting with static traditional methods in complex environments [Best viewed in color and enlarged].}
    \label{fig:preface}
    \vspace{-1.5em}
\end{figure}

Although the existing streaming approach seems promising, it still faces contradictions in real-world scenarios. These contradictions primarily stem from the diverse and unpredictable nature of driving environments. The factors such as \emph{camera motion}, \emph{weather conditions}, \emph{lighting variations}, and the presence of \emph{small objects} seriously impact the performance of perception measures, leading to fluctuations that challenge their robustness and reliability (see Sec.~\ref{sec:motivation_analysis}). This complexity in real-world scenarios underscores the limitations of a single, uniform model, which often struggles to adapt to the varied demands of different driving conditions~\cite{guo2019safe}.
In general, the challenges of streaming perception mainly include:

(1) \textit{Diverse Scenario Distribution}:~Autonomous driving environments are inherently complex and dynamic, showing a myriad of scenarios that a single perception model may not adequately address (see Fig.~\ref{fig:preface}). The need to customize perception algorithms to specific environmental conditions, while ensuring that these models operate cohesively, poses a significant challenge.~As discussed in Sec.~\ref{sec:motivation_analysis}, adapting models to various scenarios without compromising their core functionality is a crucial aspect of streaming perception.

(2) \textit{Performance-Efficiency Balance}:~To our knowledge, the integration of both large and small-scale models is essential to handle the varying complexities encountered in different driving scenes. The large models, while potentially more accurate, may suffer from increased latency, whereas smaller models may offer faster inference at the cost of reduced accuracy. Balancing performance and efficiency, therefore, becomes a challenging task. In Sec.~\ref{sec:motivation_analysis}, we explore the strategies for optimizing this balance, exploring how different model architectures can be effectively utilized to enhance streaming perception.

Generally speaking, these challenges highlight the demand for streaming perception. As we study in Sec.~\ref{sec:motivation_analysis}, addressing the \textit{diverse scenario distribution} and achieving an \textit{optimal balance between performance and efficiency} are key to advancing the state-of-the-art in autonomous driving. To address the intricate challenges presented by real-world streaming perception, we introduce \methodname, a framework designed to enhance dynamic routing capabilities in autonomous driving systems. \methodname stands as a low-rank enhanced dynamic routing framework, specifically crafted to cater to the requirements of streaming perception. It encapsulates a suite of pre-trained branch networks, each meticulously fine-tuned to optimally function under distinct environmental conditions. A key component of \methodname is the speed router module, ingeniously developed to assess and efficiently route input to the optimal branch, as detailed in Sec.~\ref{par:overview_methodname}. To sum up, the contributions are listed as:
\begin{itemize}
    \item We emphasize the impact of environmental speed as a key determinant of streaming perception. Through analysis of various environmental factors, our research highlights the imperative need for adaptive perception responsive to dynamic conditions.
    \item By utilizing a variety of streaming perception techniques, \methodname provides the speed router as a major invention. This component dynamically determines the best route for handling each input, ensuring efficiency and accuracy in perception. The ability to adapt and be versatile is demonstrated by this dynamic route-choosing mechanism.
    \item Extensive experimental evaluations have demonstrated that \methodname is capable of adapting to diverse branch selection strategies, resulting in a substantial enhancement of performance across various branch structures. This not only validates the framework's wide-ranging applicability but also confirms its effectiveness in handling different real-world scenarios.
\end{itemize}
\begin{figure*}
    \centering
    \includegraphics[width=0.9\linewidth]{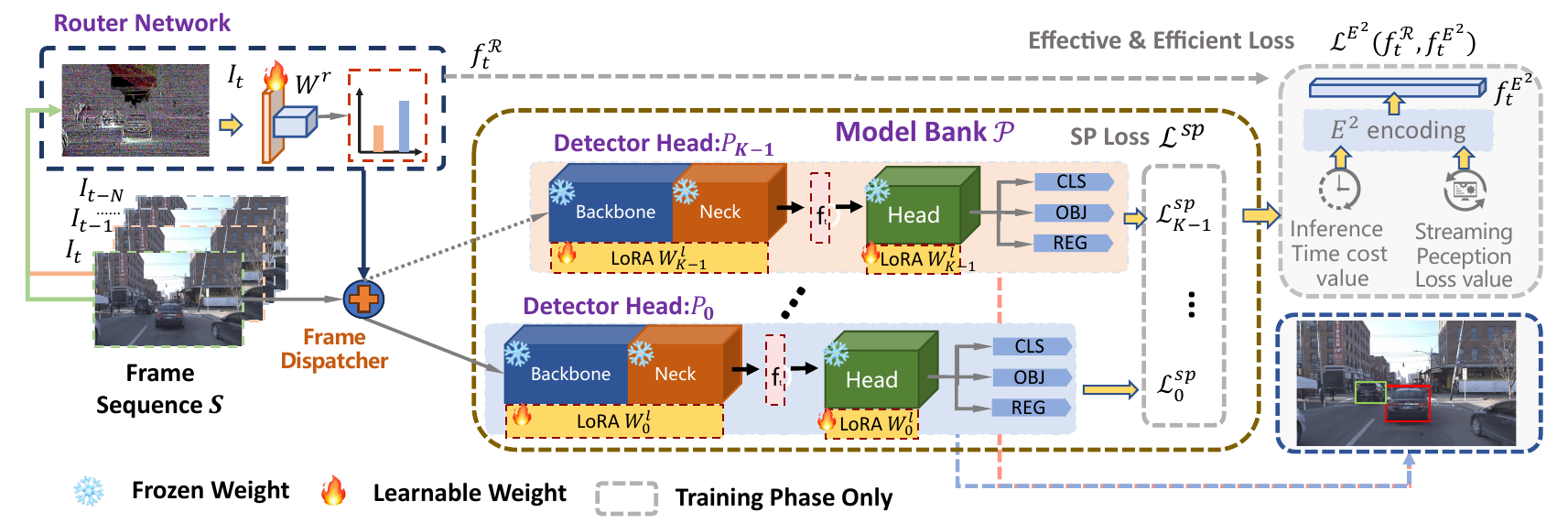}
    \vspace{-2mm}
    \caption{\small
    The \methodname Framework: This figure shows \methodname's architecture with a multi-branch network. Two branches are illustrated, each as a streaming perception sub-network. The upper right details the core architecture. Each branch processes the current frame \(I_t\) and historical frames \(I_{t-1}, I_{t-2}, \cdots, I_{t-n}\). Features are extracted by the backbone and neck, split into streams for current and historical frames, fused, then passed to the prediction head. The Speed Router selects the branch based on frame difference \(\Delta I_t\) from \(I_t\) and \(I_{t-1}\).}
    \label{fig:framework}
    \vspace{-2mm}
\end{figure*}

\vspace{-0.1in}
\section{Related Work} 
\label{sec:related_work}
This section revisits developments in streaming perception and dynamic neural networks, highlighting differences from our proposed \methodname framework. While existing methods have made progress, limitations persist in addressing real-world autonomous driving complexity.
\subsection{Streaming Perception}
\label{sub:streaming_perception}
The existing streaming perception methods fall into three main categories.~(1)~The initial methods focused on single-frame, with models like YOLOv5~\cite{yolov5} and YOLOX~\cite{ge2021yolox} achieving real-time performance.~However, lacking motion trend capture, they struggle in dynamic scenarios.~(2)~The recent approaches incorporated current and historical frames, like StreamYOLO~\cite{yang2022real} building on YOLOX with dual-flow fusion.~LongShortNet~\cite{li2023longshortnet} used longer histories and diverse fusion.~DAMO-StreamNet~\cite{he2023damo} added asymmetric distillation and deformable convolutions to improve large object perception.~(3)~Recognizing the limitations of single models, current methods explore dynamic multi-model systems.~One approach~\cite{ghosh2021adaptive} adapts models to environments via reinforcement learning.~DaDe~\cite{jo2022dade} extends StreamYOLO by calculating delays to determine frame steps.~A later version~\cite{huang2023mtd} added multi-branch prediction heads.~Beyond 2D detection, streaming perception expands into optical flow, tracking, and 3D detection, with innovations in metrics and benchmarks~\cite{wang2023estimating,sela2022context,wang2023are}.~Distinct from these existing approaches, our proposed method, \methodname, introduces a low-rank enhanced dynamic routing mechanism specifically designed for streaming perception.~\methodname stands out by integrating a suite of advanced branch networks, each fine-tuned for specific environmental conditions.~Its key innovation lies in the speed router module, which not only routes input data efficiently but also dynamically adapts to the diverse and unpredictable nature of real-world driving scenarios.

\subsection{Dynamic Neural Networks} 
\label{sub:dynamic_neural_network}
Dynamic Neural Networks (DNNs) feature adaptive network selection, outperforming static models in efficiency and performance~\cite{han2021dynamic,lan2023procontext,zhang2023improving}.~The existing research primarily focuses on structural design for core deep learning tasks like image classification~\cite{huang2018multi,h2020glance,wang2018skipnet}.~DNNs follow two approaches:~(1)~Multi-branch models~\cite{bejnordi2019international,cai2021dynamic,shazeer2017outrageously,wang2023debunking,qiao2022real,xu2022smartadapt,lee2023roma} rely on a lightweight router assessing inputs to direct them to appropriate branches, enabling tailored computation.~(2)~By generating new weights based on inputs~\cite{yang2019condconv,chen2020dynamic,su2019pixel,zhu2019deformable}, these models dynamically alter computations to match diverse needs.~DNN applications expand beyond conventional tasks.~In object detection, DynamicDet~\cite{lin2023dynamicdet} categorizes inputs and processes them through distinct branches.~This illustrates DNNs' broader applicability and potential for dynamic environments.

\section{Proposed Method}
\label{sec:method}
This section outlines the framework of our proposed \methodname.~Beginning with its underlying motivation and the critical factors driving its design, we subsequently provide an overview of its architecture and training process.

\subsection{Motivation for DyRoNet}
\label{sec:motivation_analysis}
Autonomous driving faces variability from \textit{weather}, \textit{scene complexity}, and \textit{vehicle velocity}.~By strategically analyzing key factors and routing logic, this section details the rationale behind the proposed \methodname.

\noindent \textbf{Analysis of Influential Factors}.~Statistical analysis of the Argoverse-HD dataset~\cite{li2020towards} underscores the profound influence of environmental dynamics on the effectiveness of streaming perception.~While \textit{weather} inconsistently impacts accuracy, suggesting the presence of other influential factors~(see Appendix~\red{A.1}), fluctuations in the \textit{object count} show limited correlation with performance degradation~(see Appendix~\red{A.2}).~Conversely, the presence of \textit{small objects} across various scenes poses a significant challenge for detection, especially under varying motion states~(see Appendix~\red{A.3}).~Notably, disparities in performance are most pronounced across different \textit{environmental motion states}~(see Appendix~\red{A.4}), thereby motivating the need for a dynamic, velocity-aware routing mechanism in \methodname.

\noindent \textbf{Rationale for Dynamic Routing}.~Analysis reveals that StreamYOLO's reliance on a single historical frame falters at high velocities, in contrast to multi-frame models, highlighting a connection between speed and detection performance~(see Tab.~\ref{table:comparing-sota}).~Dynamic adaptation of frame history, based on vehicular speed, enables \methodname to strike a balance between accuracy and latency~(see Sec.~\ref{sub:Comparision with SOTA}).~Through first-order differences, the system efficiently switches models to align with environmental motions.~Specifically, the dynamic routing is designed to select the optimal architecture based on the vehicle's speed profile, ensuring precision at lower velocities for detailed perception and efficiency at higher speeds for swift response.~These comprehensive analysis imforms \methodname as a robust solution for reliable perception across diverse autonomous driving scenarios.

\subsection{Architecture of DyRoNet}
\label{par:overview_methodname}
\noindent \textbf{Overview of \methodname}.~The structure of \methodname, as depicted in Fig.~\ref{fig:framework}, proposes a multi-branch structure.~Each branch within \methodname framework functions as an independent streaming perception model, capable of processing both the current and historical frames.~This dual-frame processing is central to \methodname's capability, facilitating a nuanced understanding of temporal dynamics.~Such a design is key in achieving a delicate balance between latency and accuracy, aspects crucial for real-time autonomous driving.

Mathematically, the core of \methodname lies the processing of a frame sequence, $\mathcal{S}=\{I_{t},\cdots, I_{t-N\delta t}\}$, where $N$ indicates the number of frames and $\delta t$ the interval between successive frames.~The framework process is formalized as:
\vspace{-2mm}
\begin{equation*}
\centering
\mathcal{T}=\mathcal{F}(\mathcal{S}, \mathcal{P}, \mathcal{W}),
\end{equation*}
where $\mathcal{P}=\{P_{0}, \cdots, P_{K-1}\}$ denotes a collection of streaming perception models, with each $P_{i}$ denoting an individual model within this suite.~The architecture is further enhanced by incorporating a feature extractor $\mathcal{G}_{i}$ and a perception head $\mathcal{H}_{i}$ for each model.~The Router Network, $\mathcal{R}$, is instrumental in selecting the most suitable streaming perception model for each specific scenario.

Correspondingly, the weights of \methodname are denoted by $\mathcal{W} = \{W^{d}, W^{l}, W^{r}\}$, where $W^{d}$ indicates the weights of the streaming perception model, $W^{l}$ relates to the Low-Rank Adaptation (LoRA) weights within each model, and $W^{r}$ pertains to the Router Network.~The culmination of this process is the final output, $\mathcal{T}$, a compilation of feature maps.~These maps can be further decoded through $Decode(\mathcal{T})$, revealing essential details like objects, categories, and locations.

\begin{figure}[t]
  \centering
  \includegraphics[width=0.8\linewidth]{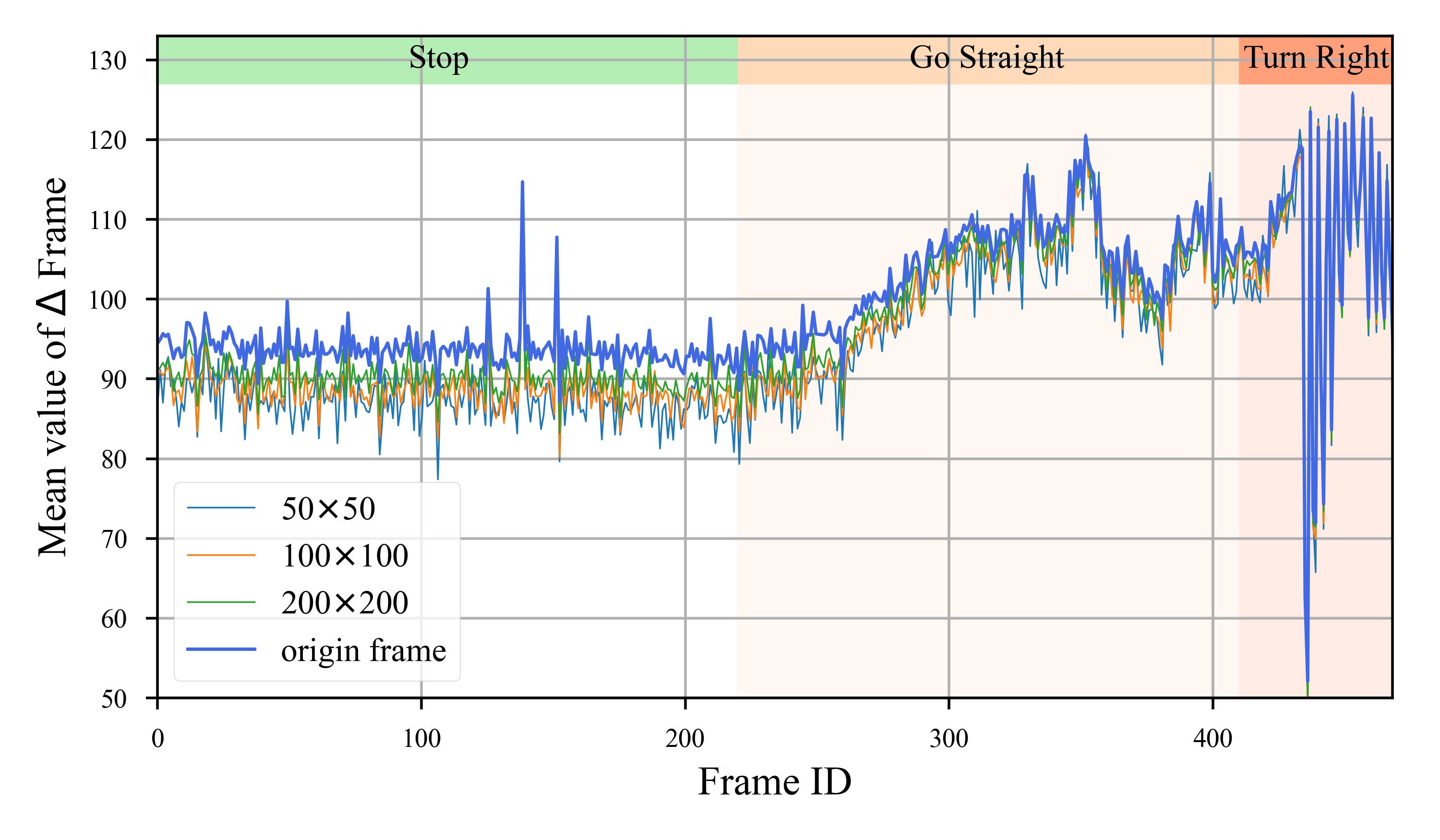}
  \caption{\small The mean curves of frame differences are depicted here.~The four curves correspond to frame sizes of the original frame, 200\(\times\)200, 100\(\times\)100, and 50\(\times\)50.~Notably, these curves show distinct fluctuations across different vehicle motion cases.}
  \label{fig:frame_diff}
  \vspace{-2.0em}
\end{figure}

\noindent \textbf{Router Network}.~The \emph{Router Network} in \methodname plays a crucial role in understanding and classifying the dynamics of the environment.~This module is designed for both environmental classification and branch decision-making.~To effectively capture environmental speed, frame differences are employed as the input to the Router Network.~As shown in Fig.~\ref{fig:frame_diff}, frame differences exhibit a high discriminative advantage for different environmental speeds.

Specifically, for frames at times \(t\) and \(t - 1\), represented as \(I_{t}\) and \(I_{t - 1}\) respectively, the frame difference is computed as $\Delta I_t = I_{t} - I_{t - 1}$.~The architecture of the Router Network, $\mathcal{R}$, is simple yet efficient.~It consists of a single convolutional layer followed by a linear layer.~The network's output, denoted as $f^{r}\in\mathbb{R}^{K}$, captures the essence of the environmental dynamics.~Based on this output, the index $\sigma$ of the optimal branch for processing the current input frame $I_{t}$ is determined through the following equation:
\vspace{-2mm}
\begin{equation}
    \sigma = \mathop{\arg\max}_K(\mathcal{R}(\Delta I_t), W^{r}),\quad\sigma\in\{0,\cdots,K-1\},
    \label{eq:optimal_index}
\end{equation}
where $\sigma$ is the index of the branch deemed most suitable for the current environmental context.~Once $\sigma$ is determined, the input frame $I_{t}$ is automatically routed to the corresponding branch by a dispatcher.

In particular, this strategy of using frame differences to gauge environmental speed is efficient.~It offers a faster alternative to traditional methods such as optical flow fields.~Moreover, it focuses on frame-level variations rather than the speed of individual objects, providing a more generalized representation of environmental dynamics.~The sparsity of $\Delta I_t$ also contributes to the robustness of this method, reducing computational complexity and making the Router Network's operations nearly negligible in the context of the overall model's performance.

\noindent \textbf{Model Bank~\&~Dispatcher}.~The core of the \methodname framework is its \textit{model bank}, which consists of an array of streaming perceptual models, denoted as $\mathcal{P}=\{{P}_{0}, \cdots, {P}_{K-1}\}$.~
Typically, the selection of the most suitable model for processing a given input is intelligently managed by the \emph{Router Network}.~This process is formalized as $P_{\sigma} = \text{Disp}(\mathcal{R}, \mathcal{P})$, where $\text{Disp}$ acts as a dispatcher, facilitating the dynamic selection of models from $\mathcal{P}$ based on the input.
The operational flow of \methodname can be mathematically defined as:
\begin{equation*}
\mathcal{T} = \mathcal{F}(\mathcal{S}, \mathcal{P}, W) = \text{Disp}(\mathcal{R}(\Delta I_{t}), \mathcal{P})(I_{t}; W^{d}_{\sigma}, W^{l}_{\sigma}),
\end{equation*}
where $\mathcal{R}$ symbolizes the \emph{Router Network}, and $\Delta I_{t}$ refers to the frame difference, a key input for model selection.~The weights $W^{d} _{\sigma}$ and $W^{l}_{\sigma}$ correspond to the selected streaming perception model and its LoRA parameters, respectively.

Note that the versatility of \methodname is further highlighted by its compatibility with a wide range of Streaming Perception models, even ones that rely solely on detectors~\cite{ge2021yolox}.
To demonstrate the efficacy of \methodname, it has been evaluated using three contemporary streaming perception models: StreamYOLO~\cite{yang2022real}, LongShortNet~\cite{li2023longshortnet}, and DAMO-StreamNet~\cite{he2023damo} (see Sec.~\ref{sub:Comparision with SOTA}).~This \textit{Model Bank~\&~Dispatcher} strategy illustrates the adaptability and robustness of \methodname across different streaming perception scenarios.

\noindent \textbf{Low-Rank Adaptation}.~A key challenge arises when fully fine-tuning individual branches, especially under the direction of \emph{Router Network}.~This strategy can lead to biases in the distribution of training data and inefficiencies in the learning process.~Specifically, lighter branches may become predisposed to simpler cases, while more complex ones might be tailored to handle intricate scenarios, thereby heightening the risk of overfitting.~Our experimental results, detailed in Sec.~\ref{sub:Comparision with SOTA}, support this observation.

To address these challenges, we have incorporated the LoRA~\cite{hu2021lora} into \methodname.~Within each model $P_{i}$, initially pre-trained on a dataset, the key components are the convolution kernel and bias matrices, symbolized as $W^{d}_{i}$.~The rank of the LoRA module is defined as $r$, a value significantly smaller than the dimensionality of $W^{d}_{i}$, to ensure efficient adaptation.~The update to the weight matrix adheres to a low-rank decomposition form, represented as 
$W_{d}^{i}+ \delta W = W_{d}^{i} + BA$.\footnote{Here, $B$ is a matrix in $R^{d\times r}$, and $A$ is in $R^{r\times k}$, ensuring that the rank $r$ remains much smaller than $d$.} This adaptation strategy allows for the original weights $W_{d}^{i}$ to remain fixed, while the low-rank components $BA$ are trained and adjusted.~The adaptation process is executed through the following projection:
\begin{equation}
    W^{d}_{i}x + \Delta Wx = W^{d}_{i}x + W^{l}_{i}x,
\end{equation}
where $x$ represents the input image or feature map, and $\Delta W=W^{l}_{i}=BA$.~The matrices $A$ and $B$ start from an initialized state and are fine-tuned during adaptation.~This approach maintains the general applicability of the model by fixing $W_{d}^{i}$, while also enabling specialization within specific sub-domains, as determined by Router Network.

In \methodname, we employ \(r=32\) for the LoRA module, though this can be adjusted based on specific requirements of the scenarios in question.~This low-rank adaptation mechanism not only enhances the flexibility of the \methodname framework but also mitigates the risk of overfitting, ensuring that each branch remains efficient and effective in its designated role.

\subsection{Training Details of DyRoNet}
\label{sec:training_details}
The training process of \methodname focuses on two primary goals:~(1)~Improving the performance of individual branches. The backpropagation only updates the chosen model's weights in this step, enabling fine-tuning on segregated samples. (2)~Achieving an optimal balance between accuracy and computational efficiency. This step only train the speed router while the remaining branches are frozen.~This dual-objective framework is represented by the overall loss function:
\begin{equation}
L = \mathcal{L}^{sp} + \mathcal{L}^{E^2},
\end{equation}
where \(\mathcal{L}^{sp}\) represents the streaming perception loss, and \(\mathcal{L}^{E^2}\) denotes the effective and efficient (E\(^2\)) loss, which supervises branch selection.

\noindent \textbf{Streaming Perception (SP) Loss}.~Each branch in \methodname is fine-tuned using its original loss function to maintain effectiveness.
The router network is trained to select the optimal branch based on efficiency supervision.~Let $\mathcal{T}_{i}=\{F_{i}^{cls}, F_{i}^{reg}, F_{i}^{obj}\}$ denote the logits produced by the \(i\)-th branch and $\mathcal{T}_{gt}=\{F_{gt}^{cls}, F_{gt}^{reg}, F_{gt}^{obj}\}$ represent the corresponding ground-truth, where \(F_{\cdot}^{cls}\), \(F_{\cdot}^{reg}\), and \(F_{\cdot}^{obj}\) are the classification, objectness, and regression logits, respectively.~The streaming perception loss for each branch, \(\mathcal{L}_{i}^{sp}\), is defined as follows:
\begin{equation}
\begin{aligned}
\mathcal{L}_{i}^{sp}(\mathcal{T}_{i}, \mathcal{T}_{gt}) =&\,\mathcal{L}_{cls}(F_{i}^{cls}, F_{gt}^{cls}) + \mathcal{L}_{obj} (F_{i}^{obj}, F_{gt}^{obj})\\
&+ \mathcal{L}_{reg}({F}_{i}^{reg}, {F}_{gt}^{reg}),
\end{aligned}
\end{equation}
where \(\mathcal{L}_{cls}(\cdot)\) and \(\mathcal{L}_{obj}(\cdot)\) are defined as Mean Square Error (MSE) loss functions, while \(\mathcal{L}_{reg}(\cdot)\) is represented by the Generalized Intersection over Union (GIoU) loss.

\noindent \textbf{Effective and Efficient (E\(^2\)) Loss}.~During the training phase, streaming perception loss values from all branches are compiled into a vector \(v^{sp} \in \mathbb{R}^{K}\), and inference time costs are aggregated into \(v^{time} \in \mathbb{R}^{K}\), with \(K\) indicating the total number of branches in \methodname.~To account for hardware variability, a normalized inference time vector \(\hat{v}^{time}=\mathrm{softmax}(v^{time})\) is introduced.~This vector is derived using the Softmax function to minimize the influence of hardware discrepancies.~The representation for effective and efficient (E\(^2\)) decision-making is defined as:
\begin{equation}
f^{E^2} = \mathcal{O}_{N}(\mathop{\arg\min}_k(\mathrm{softmax}(v^{time}) \cdot v^{sp})),
\end{equation}
where \(\mathcal{O}\) denotes one-hot encoding, producing a boolean vector of length \(K\), with the value of \(1\) at the index representing the estimated optimal branch at that moment.~The E\(^2\) Loss is then formulated as:
\begin{equation}
\mathcal{L}^{E^2}=\mathrm{KL}(f^{E^2}, f^{r}),
\end{equation}
where \(f_{r}=\mathcal{R}(\Delta I_{t})\) and \(\mathrm{KL}\) represents the Kullback-Leibler divergence, utilized to constrain the distribution.

\noindent \textbf{\methodname and relevant techniques}.~Although the structure of \methodname is somewhat similar to MoE\cite{shazeer2017outrageously,zhou2022mixture}, in contrast, the gate network of MoE is not well-suited for streaming perception.~This limitation has led to the development of speed router and the \(\mathcal{L}^{E^2}\) loss function.~\methodname, for efficiency, chooses one model over MoE's multiple experts, aligning with MoE in concept but differing in gate structure and selection strategy, making it unique for streaming contexts.

The speed router is inspired by Network Architecture Search (NAS). However, it uniquely addresses the challenge of streaming perception by transforming the search problem into an optimization of coded distances. The formulation of the loss function, \(\mathcal{L}^{E^2}\), involves converting \(v^{\text{time}}\) into a distribution via softmax, which is then combined with \(v^{\text{sp}}\) to determine the optimal model through an one-hot vector \(f^{E^2}\). This approach effectively simplifies the intricate problem of balancing accuracy and latency into a more tractable optimization task. Instead of employing NAS for loss search, our design is intricately linked to the specific needs of streaming perception, with KL divergence selected for its robustness in noisy situations\cite{kim2021comparing}. This demonstrates the efficiency and innovation of our approach.

Overall, the process of training \methodname involves striking a meticulous balance between the SP loss, which ensures the efficacy of each branch, and the E\(^2\) loss, which optimizes efficiency.~The primary objective of this training is to develop a model that not only delivers high accuracy in perception tasks but also operates within acceptable latency constraints, which is a critical requirement for real-time applications.~This balanced approach enables \methodname to adapt dynamically to varying computational resources and environmental conditions, thereby maintaining optimal performance in diverse streaming perception scenarios.

\section{Experiments}
\label{sec:experiments}

\begin{table*}[!t] \footnotesize
    \centering
    \setlength{\tabcolsep}{1.25mm}{
        \begin{tabular}{c|c|c|c|c|c|c|c|c|c|c|c}
            \toprule
Model                         & \multicolumn{2}{c|}{Random} & \multicolumn{2}{c|}{MoE} & \multicolumn{7}{c}{\methodname} \\
Bank                          & latency  & sAP & latency  & sAP & latency  & sAP & sAP$_{50}$ & sAP$_{75}$ & sAP$_{s}$ & sAP$_{m}$ & sAP$_{l}$\\
\midrule
\(\text{sYOLO}_\text{S + M}\) & 39.16 & 31.5 &  66.16 & 29.5 & \textbf{26.25} (-12.91) & \textbf{33.7} (+2.2) & 53.9 & 34.1 & 13.0 & 35.1 & 59.3 \\
\(\text{sYOLO}_\text{S + L}\) & 24.04 & 33.2 &  70.19 & 29.5 & 29.35 (+5.31) & \textbf{36.9} (+3.7)           & 58.2 & 37.5 & 14.8 & 37.4 & 64.2 \\
\(\text{sYOLO}_\text{M + L}\) & 24.69 & 35.4 &  83.65 & 33.7 & \textbf{23.51} (-1.18) & 35.0 (-0.4)           & 55.7 & 35.5 & 13.7 & 36.2 & 61.1 \\
\(\text{LSN}_\text{S + M}\)   & 24.79 & 31.8 & 128.74 & 29.8 & \textbf{21.47} (-3.32) & 30.5 (-1.3)           & 51.2 & 30.2 & 11.3 & 31.1 & 56.1 \\
\(\text{LSN}_\text{S + L}\)   & 21.49 & 33.4 & 121.62 & 29.8 & 30.48 (+8.99) & \textbf{37.1} (+3.7)           & 58.3 & 37.6 & 15.1 & 37.6 & 63.7 \\
\(\text{LSN}_\text{M + L}\)   & 24.75 & 35.6 & 136.66 & 34.1 & 29.05 (+4.30) & \textbf{36.9} (+1.3)           & 58.2 & 37.4 & 14.9 & 37.5 & 63.3 \\
\(\text{DAMO}_\text{S + M}\)  & 36.61 & 33.5 & 188.42 & 31.8 & \textbf{33.22} (-3.39) & \textbf{35.5} (+2.0)  & 56.9 & 36.2 & 14.4 & 36.8 & 63.2 \\
\(\text{DAMO}_\text{S + L}\)  & 35.12 & 34.5 & 188.57 & 31.8 & 39.60 (+4.48) & \textbf{37.8} (+3.3)           & 59.1 & 38.7 & 16.1 & 39.0 & 64.2 \\
\(\text{DAMO}_\text{M + L}\)  & 37.30 & 36.5 & 195.87 & 35.5 & 37.61 (+0.31) & \textbf{37.8} (+1.3)           & 58.8 & 38.8 & 16.1 & 39.0 & 64.0 \\
            \bottomrule
    \end{tabular}}
    \caption{\small Comparison of latency (ms) and the corresponding sAP performance on 1x RTX 3090.~The values are highlighted in \textbf{bold} font if the \methodname perform better than the corresponding \emph{random} and \emph{MoE} case. Due to the overall poorer performance of \emph{MoE}, we only lists the relative latency and sAP differences between \methodname and the \emph{random} approach.}
    \label{table:inference_time}
    \vspace{-1.5em}
\end{table*}

\begin{table}[!t] \footnotesize
    \centering
    \setlength{\tabcolsep}{1.25mm}{
        \begin{tabular}{c|c|c}
            \toprule
            Methods   & Latency (ms) & sAP $\uparrow$ \\
            \midrule
            \multicolumn{3}{ c }{Non-real-time detector-based methods} \\
            \midrule
            Adaptive Streamer \cite{ghosh2021adaptive}  & - & 21.3\\
            Streamer (S=600) \cite{li2020towards}       & - & 20.4\\
            Streamer (S=900) \cite{li2020towards}       & - & 18.2\\
            Streamer+AdaScale \cite{ghosh2021adaptive}  & - & 13.8\\
            \midrule
            \multicolumn{3}{ c }{Real-time detector-based methods}\\
            \midrule
            DAMO-StreamNet-L \cite{he2023damo}          & 39.6 & 37.8\\
            LongShortNet-L \cite{li2023longshortnet}    & 29.9 & 37.1\\
            StreamYOLO-L \cite{yang2022real}            & 29.3 & 36.1\\
            DAMO-StreamNet-M \cite{he2023damo}          & 33.5 & 35.7\\
            LongShortNet-M \cite{li2023longshortnet}    & 25.1 & 34.1\\
            StreamYOLO-M \cite{yang2022real}            & 24.8 & 32.9\\
            DAMO-StreamNet-S \cite{he2023damo}          & 30.1 & 31.8\\
            LongShortNet-S \cite{li2023longshortnet}    & 20.3 & 29.8\\
            StreamYOLO-S \cite{yang2022real}            & 21.3 & 28.8\\
            \midrule
            \methodname (\(\text{DAMO}_\text{M + L}\))    & \underline{37.61} (-1.99) & \textbf{37.8} (same) \\
            \methodname (\(\text{LSN}_\text{M + L}\))     & \underline{29.05} (-0.85) & 36.9 (-0.2) \\
            \methodname (\(\text{sYOLO}_\text{M + L}\))   & \underline{23.51} (-5.79) & 35.0 (-1.1) \\
            \bottomrule
    \end{tabular}}
    \caption{\small The comparison of \methodname and SOTA.~The optimal values over its larger model are highlighted in \textbf{bold} font and the optimal values of online evaluation latency are shown in \underline{underline} font. The latency of \methodname is evaluated on 1x RTX 3090 and compared with the latency of corresponding smaller model.}
    \label{table:comparing-sota}
    \vspace{-1.5em}
\end{table}

\subsection{Dataset and Metric}%
\label{sub:Dataset and Metric}
\noindent \textbf{Dataset}.~Argoverse-HD dataset~\cite{li2020towards} is utilized for our experiments, specifically designed for streaming perception in autonomous driving scenarios.~It comprises high-resolution RGB images captured from urban city street, offering a realistic representation of diverse driving conditions.~The dataset is structured into two main segments: a training set consisting of 65 video clips and a test set comprising 24 video clips.~Each video clip in the dataset spans over 600 frames in average, contributing to a training set with approximately 39k frames and a test set containing around 15k frames.~Notably, Argoverse-HD provides high-frame-rate (30fps) 2D object detection annotations, ensuring accuracy and reliability without relying on interpolated data.

\noindent \textbf{Evaluation Metric}.~Streaming Average Precision (sAP) are adopted as the primary metric for performance evaluation, which is widely recognized for its effectiveness in streaming perception tasks \cite{li2020towards}. It offers a comprehensive assessment by calculating the mean Average Precision (mAP) across various Intersection over Union (IoU) thresholds, ranging from 0.5 to 0.95.~This metric allows us to evaluate detection performance across different object sizes, including large, medium, and small objects, providing a robust measurement in real-world streaming perception scenarios.

\subsection{Implementation Details}
\label{sub:Implementation Details}
We tested three state-of-the-art streaming perception models: StreamYOLO\cite{yang2022real}, LongShortNet\cite{li2023longshortnet}, and DAMO-StreamNet\cite{he2023damo}.~These models, integral to the \methodname architecture, come with pre-trained parameters across three distinct scales: small (S), medium (M), and large (L), catering to a variety of processing requirements.~In constructing the model bank \(\mathcal{P}\) for \methodname, we strategically selected different model configurations to evaluate performance across diverse scenarios.~For instance, the notation \methodname(\(\text{DAMO}_\text{S + M}\)) represents a configuration where \methodname employs the small (S) and medium (M) scales of DAMO-StreamNet as its two branches.\footnote{Similar notations are used for other model combinations, allowing for a systematic exploration of the framework's adaptability and performance under varying computational constraints.}~All experiments were conducted on a high-performance computing platform equipped with Nvidia 3090Ti GPUs (x4), ensuring robust and reliable computational power to handle the intensive processing demands of the streaming perception models.~This setup provided a consistent and controlled environment for evaluating the efficacy of \methodname across different model configurations, contributing to the thoroughness and validity of our results. For more implementation details, please refer to Appendix~\red{C}.

\subsection{Comparision with SOTA Methods}%
\label{sub:Comparision with SOTA}
We compared \methodname with state-of-the-art methods to evaluate its performance.~In this subsection, we directly copied the reported sAP performance from their original papers as their results, for fair comparison, we evaluated the latency of each real-time model on 1x RTX 3090.~The performance comparison was conducted on the Argoverse-HD dataset~\cite{li2020towards}.~An overview of the results reveals that our proposed \methodname with a model bank of DAMO-StreamNet series achieves 37.8\% sAP in 37.61 ms latency, outperforming the current state-of-the-art methods in latency by a significant margin.~This demonstrates the effectiveness of the systematic improvements in \methodname.

\begin{table}[!t] \footnotesize
    \centering
    \setlength{\tabcolsep}{1.25mm}{
        \begin{tabular}{c|c c c|c c}
            \toprule
            Model Bank                              & Full (sAP) & LoRA (sAP) \\
            \midrule
            \(\text{StreamYOLO}_\text{S + M}\)      & 32.9 & \textbf{33.7} (+0.8) \\
            \(\text{StreamYOLO}_\text{S + L}\)      & 36.1 & \textbf{36.9} (+0.8) \\
            \(\text{StreamYOLO}_\text{M + L}\)      & \textbf{36.2} & 35.0 (-1.2) \\
            \(\text{LongShortNet}_\text{S + M}\)    & 29.0 & \textbf{30.5} (+1.5) \\
            \(\text{LongShortNet}_\text{S + L}\)    & 36.2 & \textbf{37.1} (+0.9) \\
            \(\text{LongShortNet}_\text{M + L}\)    & 36.3 & \textbf{36.9} (+0.6) \\
            \(\text{DAMO-StreamNet}_\text{S + M}\)  & 34.8 & \textbf{35.5} (+0.7) \\
            \(\text{DAMO-StreamNet}_\text{S + L}\)  & 31.1 & \textbf{37.8} (+6.7) \\
            \(\text{DAMO-StreamNet}_\text{M + L}\)  & 37.4 & \textbf{37.8} (+0.4) \\
            \bottomrule
    \end{tabular}}
    \caption{\small Comparison of LoRA finetune and Full finetune.~The optimal values between \emph{Full} and \emph{LoRA} are shown in \textbf{bold} font.}
    \label{table:finetune_compare}
    \vspace{-2em}
\end{table}

\begin{table}[!t] \footnotesize
    \centering
    \setlength{\tabcolsep}{1.25mm}{
        \begin{tabular}{c|c|c c c c}
            \toprule
                                   & Model Bank                       & \(b_0\) & \(b_1\) & \(b_2\) & sAP \\
            \midrule
            \multirow{9}{*}{\shortstack{\(K=2\)\\same\\model}} & \(\text{DAMO}_\text{S + M}\)  & 31.8 & 35.5 & - & 35.5 \\
                                   & \(\text{DAMO}_\text{S + L}\)  & 31.8 & 37.8 & - & 37.8 \\
                                   & \(\text{DAMO}_\text{M + L}\)  & 35.5 & 37.8 & - & 37.8 \\
                                   & \(\text{LSN}_\text{S + M}\)   & 29.8 & 34.1 & - & 30.5 \\
                                   & \(\text{LSN}_\text{S + L}\)   & 29.8 & 37.1 & - & 37.1 \\
                                   & \(\text{LSN}_\text{M + L}\)   & 34.1 & 37.1 & - & 36.9 \\
                                   & \(\text{sYOLO}_\text{S + M}\) & 29.5 & 33.7 & - & 33.7 \\
                                   & \(\text{sYOLO}_\text{S + L}\) & 29.5 & 36.9 & - & 36.9 \\
                                   & \(\text{sYOLO}_\text{M + L}\) & 33.7 & 36.9 & - & 35.0 \\
            \midrule
            \multirow{5}{*}{\shortstack{\(K=2\)\\different\\model}} & \(\text{DAMO}_\text{S}\) + \(\text{LSN}_\text{S}\)    & 31.8 & 29.8 & - & 30.5 \\
                                   & \(\text{DAMO}_\text{S}\) + \(\text{LSN}_\text{M}\)    & 31.8 & 34.1 & - & 34.1 \\
                                   & \(\text{DAMO}_\text{S}\) + \(\text{LSN}_\text{L}\)    & 31.8 & 37.1 & - & 31.8 \\
                                   & \(\text{DAMO}_\text{M}\) + \(\text{LSN}_\text{S}\)    & 35.5 & 29.8 & - & 29.8 \\
                                   & \(\text{DAMO}_\text{L}\) + \(\text{LSN}_\text{S}\)    & 37.8 & 29.8 & - & 29.8 \\
            \midrule
            \multirow{3}{*}{\shortstack{\(K=3\)\\same\\model}} & \(\text{DAMO}_\text{S + M + L}\)                      & 31.8 & 35.5 & 37.8 & 37.7 \\
                                   & \(\text{LSN}_\text{S + M + L}\)                       & 29.8 & 34.1 & 37.1 & 36.1 \\
                                   & \(\text{sYOLO}_\text{S + M + L}\)                     & 29.5 & 33.7 & 36.9 & 36.6 \\
            \bottomrule
    \end{tabular}}
    \caption{\small The performance of \methodname under various model bank selection.~\(K\) means the number of the model in bank \(\mathcal{P}\).}
    \label{table:branch_compare}
    \vspace{-2em}
\end{table}

\subsection{Inference Time}
\label{sub:inference_time}

In this subsection, we conducted detailed experiments analyzing the trade-offs between \methodname's inference time and performance under different model bank selection.
It is notable that the latency of \methodname presented in Tab.~\ref{table:comparing-sota} do not accurately reflect the real outperform.~Due to the varying hardware platforms for measuring latency across methods, a fair comparison cannot be achieved.~For instance, DAMO-StreamNet is tested on 1x V100.~To address these differences, we conducted additional tests on a 1x RTX 3090, which highlight DyRoNet's performance enhancements.~Tab.~\ref{table:inference_time} presents the latency comparison and highlights \methodname's superior performance—maintaining competitive inference speed alongside accuracy gains versus the \emph{random} and \emph{MoE} approaches.~Where the \emph{MoE} predicts the weights of each branch via a gate module and then combines the output results accordingly.
Specifically, \methodname achieves efficient speeds while preserving or enhancing performance.~This balance enables meeting real-time needs without compromising perception quality, critical for autonomous driving where both factors are paramount.~By validating effectiveness in inference time reductions and accuracy improvements, the results show the practicality and efficiency of \methodname's dynamic model selection.

\subsection{Ablation Study}
\label{sub:Ablation Study}

\noindent \textbf{Router Network.}~To validate the effectiveness of the \emph{Router Network} based on frame difference, we conducted comparative experiments using frame difference \(\Delta I_t\), the current frame \(I_t\), and the concatenation of the consecutive frames \([I_t + I_{t - 1}]\) as input modality of the \emph{Router Network}.~
For comparison, a naive method of input guidance also employed: By calculating \(\mathbb{E}(\Delta I_t)\), the larger branch is selected if \(\mathbb{E}(\Delta I_t) > 0\), otherwise the smaller branch is chosen.
~The results are presented in Tab.~\ref{table:router_input}.~For comparison, the \emph{Router Network} only be trained in these experiments while the leftover parts be frozen.

It shows that using \(\Delta I_t\) as input exhibits better performance than other methods (35.0 sAP of \(\text{sYOLO}_\text{S + L}\) and 34.6 sAP of \(\text{sYOLO}_\text{M+L}\)).
This indicates that utilizing \(\Delta I_t\) offers significant advantages in comprehending and characterizing environmental speed.~Conversely, it also underscores that employing single frames as input or using multiple frames as input renders the lightweight model bank selection model ineffective.
Furthermore, the proportion of sample splits across branches can also illustrate the discriminative power with respect to environmental factors.
For instance, the \(\mathbb{E}(\Delta I_t)\) criterion resulting in a evenly spliting distribution (48.22\% \(\mathbb{E}(\Delta I_t) > 0\) over train set and 49.85\% \(\mathbb{E}(\Delta I_t) > 0\) over test set).~Indicating the direct sample selection without router lacks estimation of environmental factors, thereby weakening its discriminative power.

In contrast, Tab.\ref{tab:2} presents statistics indicating the router layer's effectiveness in allocating samples to specific models and showcase its ability to strike a balance between latency and performance.~This balance is crucial for streaming perception and underscores our contribution.

\begin{table}[!t] \footnotesize
    \begin{center}
        \footnotesize
        \begin{tabular}[c]{l|r|r|r|r} 
            \toprule
            Model & \multicolumn{2}{c|}{training time} & \multicolumn{2}{c}{inference time}\\
            Combination & Model 1 & Model 2 & Model 1 & Model 2\\
            \midrule
            SYOLO (M+L) & 37.53\% & 62.47\% & 94.67\% &   5.33\% \\
            LSN (M+L)   & 30.86\% & 69.14\% & 19.87\% &  80.13\% \\
            DAMO (M+L)  & 84.61\% & 15.39\% &  0.02\% &  99.98\% \\
            \bottomrule
        \end{tabular}
        \vspace{-1em}
    \end{center}
    \caption{The statistics of model selection by \methodname under different model choices during both training and inference time.}\label{tab:2}
    \vspace{-1em}
\end{table}

\begin{table}[!t]\footnotesize
    \centering
    \setlength{\tabcolsep}{1.25mm}{
        \begin{tabular}{c|c|c|c|c}
            \toprule
            Model & \multicolumn{4}{c}{Input Modality / Criterion} \\
            Bank &  $I_{t}$ & $[I_{t}+I_{t-1}]$ & $\mathbb{E}(\Delta I_t)$ & $\Delta I_{t}$ (\methodname)\\
            \midrule
            \(\text{sYOLO}_\text{S + M}\) & \textbf{33.7} & \textbf{33.7} & 31.5 & 32.6\\
            \(\text{sYOLO}_\text{S + L}\) & 34.1          & 30.2          & 32.9 & \textbf{35.0}\\
            \(\text{sYOLO}_\text{M + L}\) & 33.7          & 33.7          & 34.2 & \textbf{34.6}\\
            \bottomrule
    \end{tabular}}
    \caption{\small Ablation of router network input / criterion.~The optimal results are marked in \textbf{bold} font under the same model bank setting.}
    \label{table:router_input}
    \vspace{-2em}
\end{table}

\noindent \textbf{Branch Selection.}
Our research on streaming perception models has shown that configuring these models across varying scales can optimize their performance.~We found that combining L and S models strikes an optimal balance, resulting in significant speed improvements.~This conclusion is supported by the empirical evidence presented in Tab.~\ref{table:branch_compare}, which clearly shows that the L+S model pairing outperforms both the L+S and L+M cases.~Our findings highlight the importance of strategic model scaling in streaming perception and provide a framework for future model optimization in similar domains.

\noindent \textbf{Fine-tuning Scheme.} 
We contrasted the performance of direct fine-tuning with the LoRA fine-tuning strategy~\cite{zhu2023tracking} for streaming perception models.
~Tab.~\ref{table:finetune_compare} shows that LoRA fine-tuning surpasses direct fine-tuning, with the DAMO-Streamnet-based model bank configuration realizing an absolute gain of over 1.6\%.~This substantiates LoRA's fine-tuning proficiency in circumventing the pitfalls of forgetting and data distribution bias inherent to direct fine-tuning.
This result demonstrates that LoRA fine-tuning can effectively mitigate the overfitting while fine-tuning, leading to a stable performance improvement.

\noindent \textbf{LoRA Rank.} To assess the impact of LoRA ranks in \methodname, we conducted experiments with rank \(r = 32, 16, 8\) respectively.~All experiments were train for 5 epochs between \emph{Router Network} training and model bank fine-tuning.~The results are presented in Tab.~\ref{table:lora_rank}.~It can be observed that the performance is better with \(r=32\) compared to \(r=8\) and \(r=16\), and only occupy 10\% of the total model parameters.~Therefore, based on these experiments, \(r=32\) was selected as the default setting for our experiments.
Although a smaller LoRA rank occupies fewer parameters, it leads to a rapid performance decay.~The experimental results clearly demonstrate that with LoRA fine-tuning, it is possible to achieve superior performance than a single model while utilizing a smaller parameter footprint.

\begin{table}[!t] \footnotesize
    \centering
    \setlength{\tabcolsep}{1.25mm}{
        \begin{tabular}{c|c| c c|c r}
            \toprule
            Model Bank & Rank & branch 0 & branch 1 & after train & Param.(\%) \\
            \midrule
            \(\text{DAMO}_{\text{S + L}}\)  &  8 & 31.8 & 37.8 & 35.9         &  4.02 \\
            \(\text{DAMO}_{\text{S + L}}\)  & 16 & 31.8 & 37.8 & 35.9         &  7.73 \\
            \(\text{DAMO}_{\text{S + L}}\)  & 32 & 31.8 & 37.8 & \textbf{37.8} & 14.35 \\
            \(\text{LSN}_{\text{S + L}}\)   &  8 & 29.8 & 37.1 & 30.6         &  5.48 \\
            \(\text{LSN}_{\text{S + L}}\)   & 16 & 29.8 & 37.1 & 30.6         &  5.48 \\
            \(\text{LSN}_{\text{S + L}}\)   & 32 & 29.8 & 37.1 & \textbf{36.9} & 10.39 \\
            \(\text{sYOLO}_{\text{S + L}}\) &  8 & 29.5 & 36.9 & 35.0         &  2.7 \\
            \(\text{sYOLO}_{\text{S + L}}\) & 16 & 29.5 & 36.9 & 35.0         &  5.38 \\
            \(\text{sYOLO}_{\text{S + L}}\) & 32 & 29.5 & 36.9 & \textbf{36.6} & 10.21 \\
            \bottomrule
    \end{tabular}}
    \caption{\small Ablation of LoRA rank: In the \emph{Param.} column, we solely compare the proportion of parameters occupied by LoRA to the entire model.~The best performance under the same model bank setting are highlighted in \textbf{bold} font.}
    \label{table:lora_rank}
    \vspace{-1em}
\end{table}

\subsection{Extra Experiment on NuScenes-H dataset}%
\label{sub:Extra Experiment on NuScenes-H dataset}
To validate the \methodname on other dataset, we converted the 3D streaming perception dataset nuScenes-H\cite{wang2023are} into 2D format.~The experiment details are provided in the Appendix \refred{D}.~As shown in Tab.~\ref{table:nuscenes_exp}, \methodname consistently achieves better results than other branch fusion methods on nuScenes-H 2D dataset.~It demonstrates \methodname's advantages in branch fusion selection and its versatility.

\begin{table}[!t] \footnotesize
    \centering
    \setlength{\tabcolsep}{1.25mm}{
        \begin{tabular}{c|c c|c c c}
            \toprule
            Model Bank & b1 & b2 & Random & MoE & DyRoNet \\
            \midrule
            \(\text{sYOLO}_{\text{S + M}}\)                      & 6.9 & 9.2 & 8.1 & 6.9 & \textbf{8.9}\\
            \(\text{sYOLO}_{\text{S + L}}\)                      & 7.3 & 9.9 & 8.6 & 7.3 & \textbf{9.0}\\
            \(\text{sYOLO}_{\text{M + L}}\)                      & 8.9 & 9.6 & 9.2 & 8.9 & \textbf{9.3}\\
            \(\text{sYOLO}_{\text{S}} + \text{LSN}_{\text{S}}\) & 6.6 & 6.2 & 6.4 & 6.2 & \textbf{6.5}\\
            \(\text{sYOLO}_{\text{M}} + \text{LSN}_{\text{M}}\) & 9.1 & 9.3 & 9.2 & 9.1 & \textbf{9.3}\\
            \(\text{sYOLO}_{\text{L}} + \text{LSN}_{\text{L}}\) & 9.0 & 9.6 & 9.3 & 9.0 & \textbf{9.6}\\
            \bottomrule
    \end{tabular}}
    \caption{\small The sAP results are shown in the table.~Where \emph{b1} denotes the independent performance of the first branch and \emph{b2} denotes the second one.~The branch fusion method \emph{Random} and \emph{MoE} are similar with Tab.~\ref{table:inference_time}.~The best method is highlighted in \textbf{bold} font.}
    \label{table:nuscenes_exp}
    \vspace{-2em}
\end{table}

\vspace{-2mm}
\section{Conclusion}
\label{sec:conclusion}
In conclusion, we present the \textbf{Dy}namic \textbf{Ro}utering \textbf{Net}work (\methodname), a system that dynamically selects specialized detectors for varied environmental conditions with minimal computational overhead.
Our Low-Rank Adapter mitigates distribution bias, thereby enhancing scene-specific performance.
Experimental results validate \methodname's state-of-the-art performance, offering a benchmark for streaming perception and insights for future research.~In the future, \methodname's principles will inform the development of more advanced, reliable systems.

\vspace{-2mm}
\section*{Acknowledgments}
Zhi-Qi Cheng's research in this project was supported in part by the US Department of Transportation, Office of the Assistant Secretary for Research and Technology, under the University Transportation Center Program (Federal Grant Number 69A3551747111), and Intel and IBM Fellowships.

{\small
\bibliographystyle{ieee_fullname}
\bibliography{egbib}
}

\appendix
\clearpage
\twocolumn[
  \begin{@twocolumnfalse}
    \section*{\centering \Large DyRoNet: Dynamic Routing and Low-Rank Adapters for Autonomous Driving Streaming Perception\\(Supplementary Material)}
    \vspace{10em} %
  \end{@twocolumnfalse}
\vspace{-25mm}
]

\renewcommand\thesection{\Alph{section}}
\setcounter{section}{0}

\noindent The appendix completes the main paper by providing in-depth research details and extended experimental results.~The structure of the appendix is organized as follows:
    \begin{enumerate}
        \item Analysis of Environmental Factors Affecting Streaming Perception: Sec.~\ref{sec:factor_analysis}
            \begin{itemize}
                \item Impact of Weather Conditions: Sec.~\ref{sub:weather}
                \item Quantitative Analysis of Objects: Sec.~\ref{sub:The quantity of objects}
                \item Proportion of Small Objects: Sec.~\ref{sub:The quantity of small objects}
                \item Environmental Speed Dynamics: Sec.~\ref{sub:The speed of environment}
            \end{itemize}
        \item Expanded Experimental Results: Sec.~\ref{sec:More experiment results}
            \begin{itemize}
                \item Inference Time: Analysis Sec.~\ref{sub:Inference Time}
                \item Statistic of model selection: Sec.~\ref{sub:Statistic of model selection}
                \item The comparison between Speed Router and \(\mathbb{E}[\Delta I_t]\): Sec.~\ref{sub:The comparison between Speed Router and mean of delta I_t}
            \end{itemize}
        \item Detailed Description of \methodname: Sec.~\ref{sec:More implementation details}
            \begin{itemize}
                \item Selection of Pre-trained Model: Sec.~\ref{sub:Pre-trained model selection}
                \item Hyperparameter Settings: Sec.~\ref{sub:hyperparameters}
            \end{itemize}
        \item Detailed Description of Experiments on nuScenes-H Dataset: Sec.~\ref{sub:experiments_on_nuscenes_h}
    \end{enumerate}

\section{Factor Analysis in Streaming Perception}
\label{sec:factor_analysis}
In development of \methodname, we undertook an extensive survey and analysis to identify key influencing factors in autonomous driving scenarios that could potentially impact streaming perception.~This analysis utilized the Argoverse-HD dataset~\cite{li2020towards}, a benchmark in the field of streaming perception.~The primary goal of this factor analysis was to isolate the most critical factor affecting streaming perception performance.~As elaborated in the main text, our comprehensive analysis led to the identification of the speed of the environment as the predominant factor.~Consequently, \methodname is tailored to address this specific aspect.~Our analysis focuses on four primary elements: \emph{weather conditions}, \emph{object quantity}, \emph{small object proportion}, and \emph{environmental speed}.~We methodically examined each of these factors to evaluate their respective impacts on streaming perception within autonomous driving.

\begin{figure}[!t]
    \centering
    \subfloat[]{%
        \includegraphics[width=.31\linewidth]{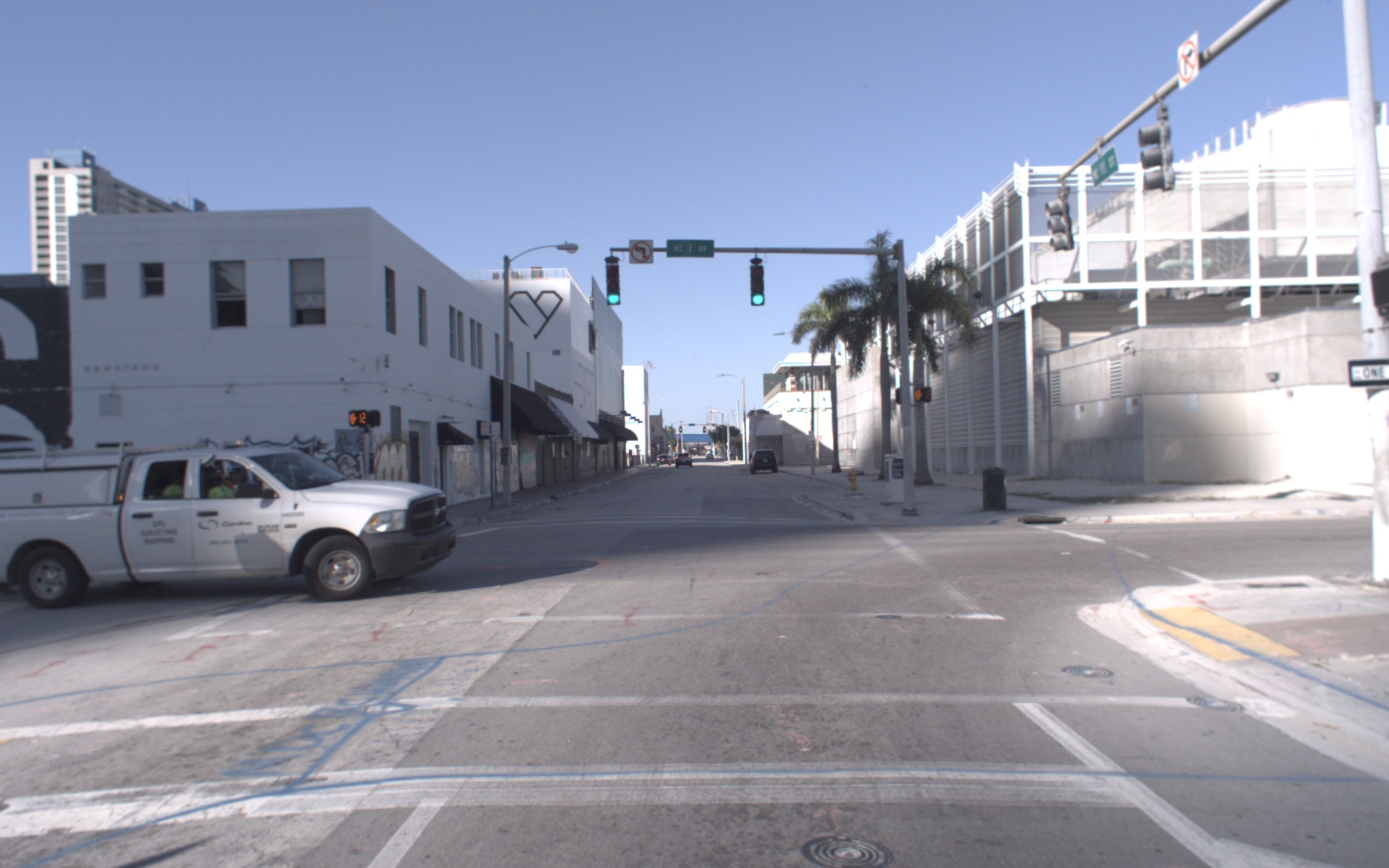}
    }
    \subfloat[]{%
        \includegraphics[width=.31\linewidth]{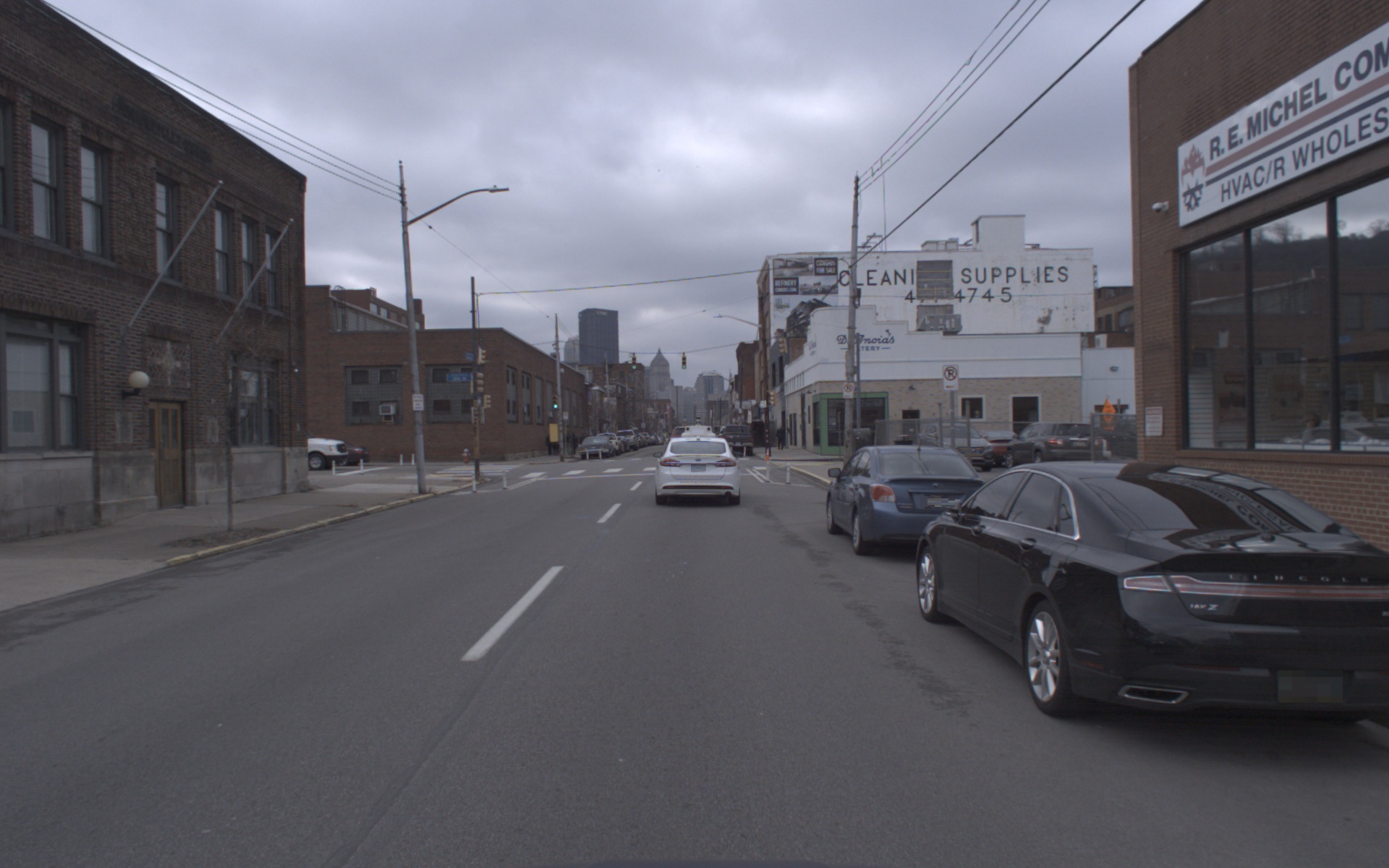}
    }\\
    \subfloat[]{%
        \includegraphics[width=.31\linewidth]{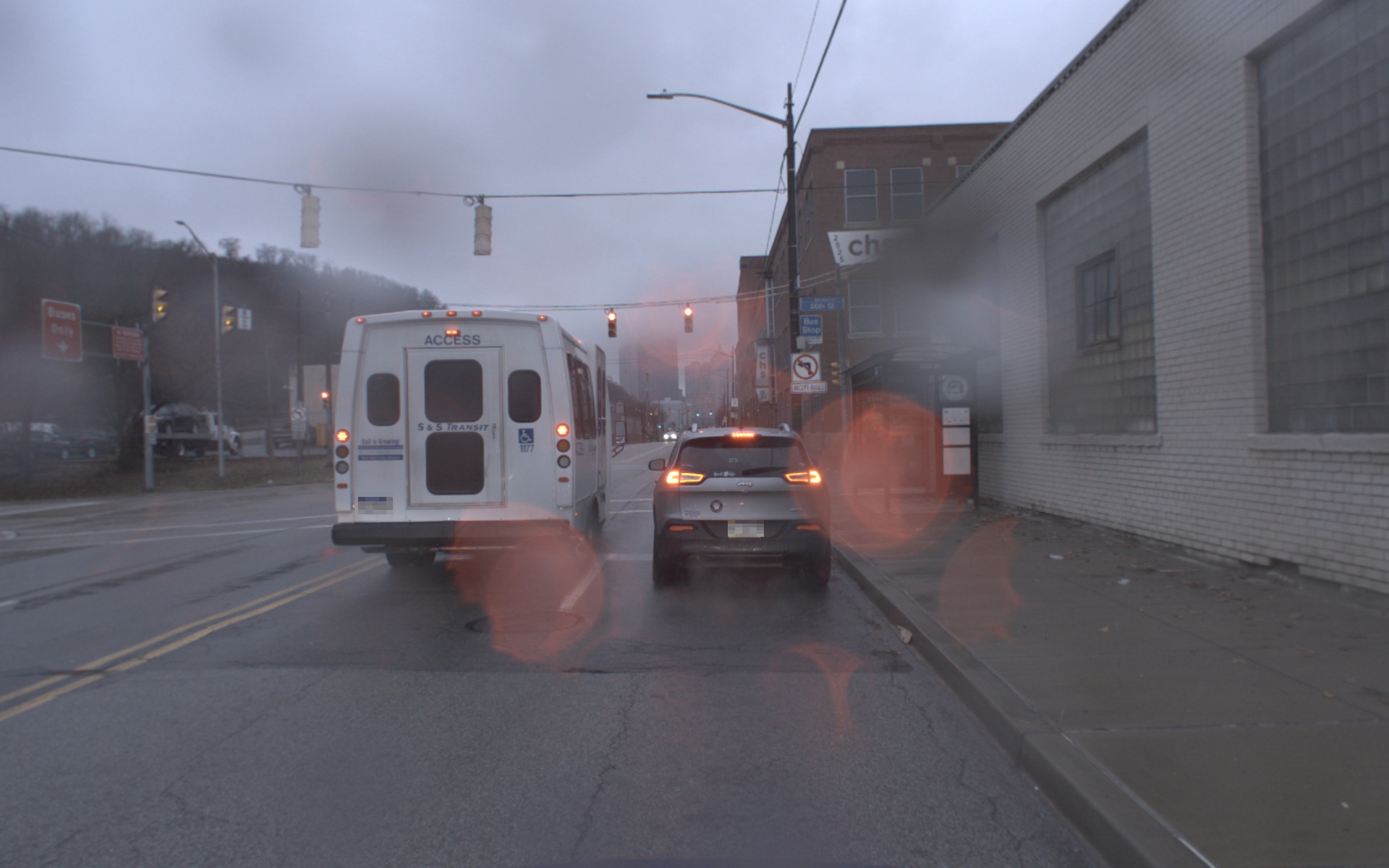}
    }
    \subfloat[]{%
        \includegraphics[width=.31\linewidth]{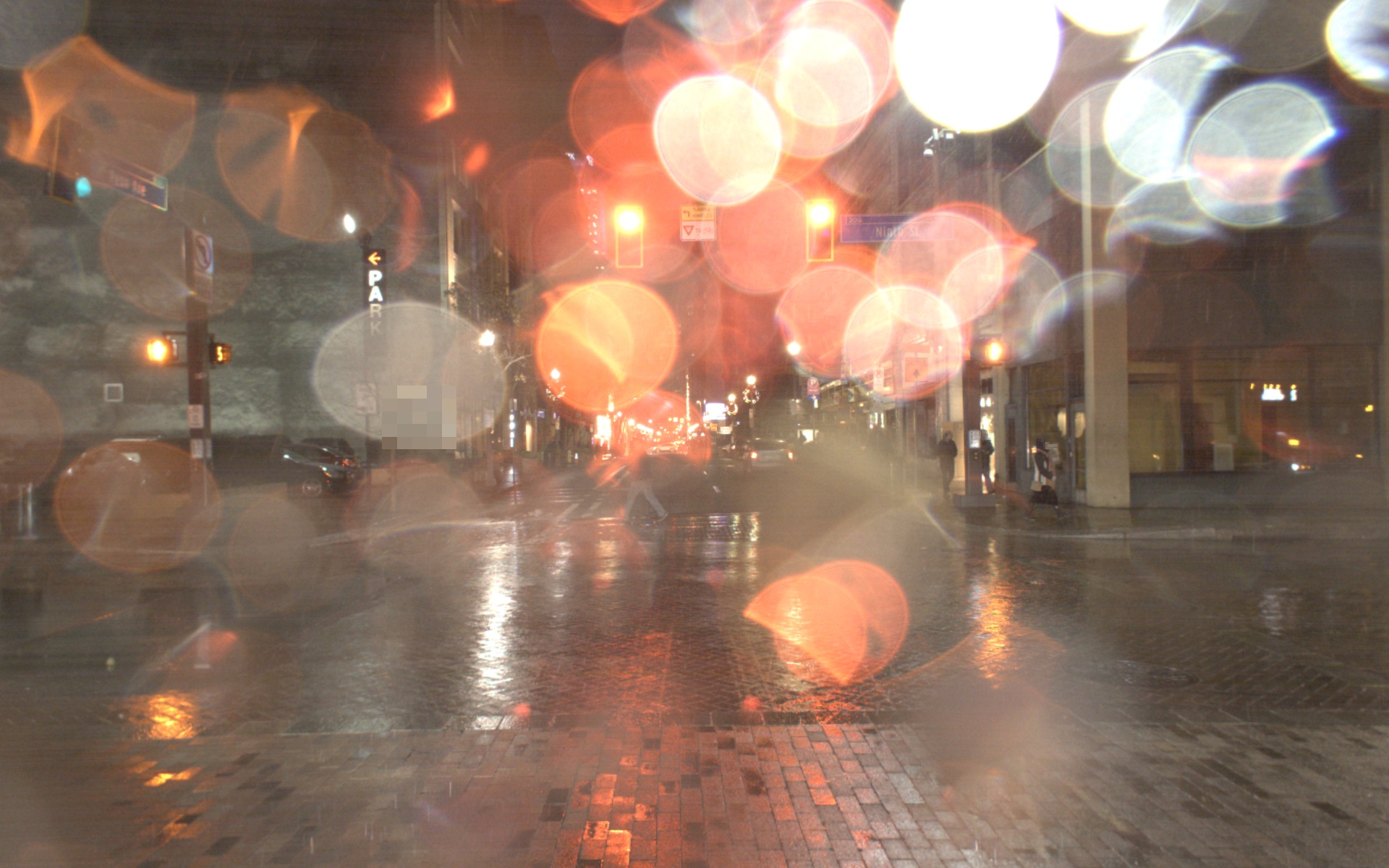}
    }
    \subfloat[]{%
        \includegraphics[width=.31\linewidth]{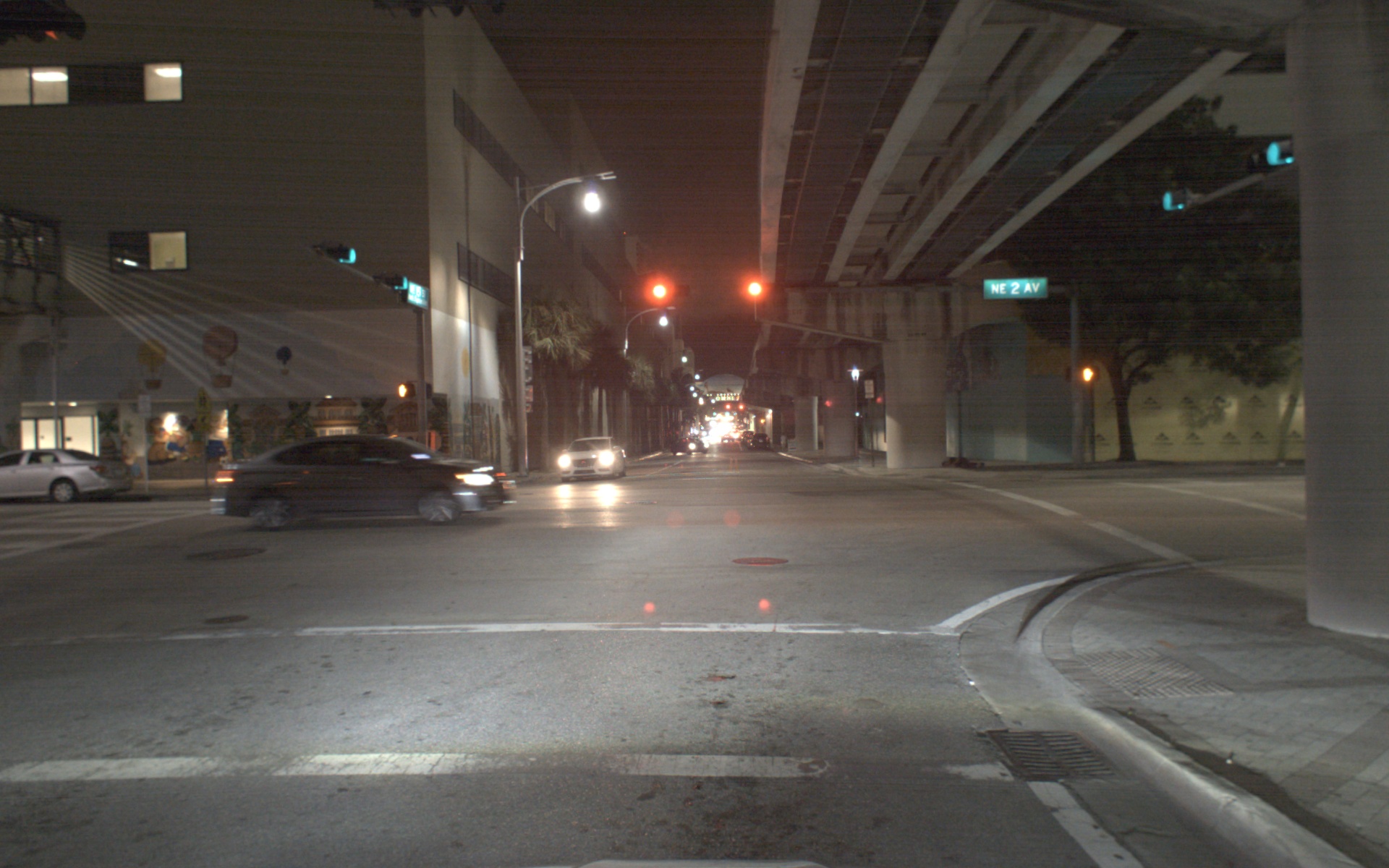}
    }
    \caption{\small Illustrative Examples of Varied Weather Conditions and Times of Day: (a) Sunny during Daytime, (b) Cloudy during Daytime, (c) Rainy during Daytime, (d) Rainy during Nighttime, (e) Sunny during Nighttime.}
    \label{fig:appendix_1}
\end{figure}

\subsection{Impact of Weather Conditions}%
\label{sub:weather}
The Argoverse-HD dataset, comprising testing, training, and validation sets, includes a diverse range of weather conditions.~Specifically, the dataset contains 24, 65, and 24 video segments in the testing, training, and validation sets, respectively, with frame counts ranging from 400 to 900 per segment.~Tab.~\ref{table:appendix_1} details the distribution of various weather types across these subsets.~Fig.~\ref{fig:appendix_1} provides visual examples of different weather conditions captured in the dataset.~A clear variation in visual clarity and perception difficulty is observable under different conditions, with scenarios like Sunny + Day or Cloudy + Day appearing visually more challenging compared to Rainy + Night.

To evaluate the impact of weather conditions on streaming perception, we conducted tests using a range of pre-trained models from StreamYOLO~\cite{yang2022real}, LongShortNet~\cite{li2023longshortnet}, and DAMO-StreamNet~\cite{he2023damo}, employing various scales and settings.~The results, presented in Tab.~\ref{table:appendix_2}, indicate that performance is generally better during Day conditions compared to Night.~This confirms that weather conditions indeed influence streaming perception.

\begin{figure}[ht]
    \centering
    \subfloat[]{%
        \includegraphics[width=.45\linewidth]{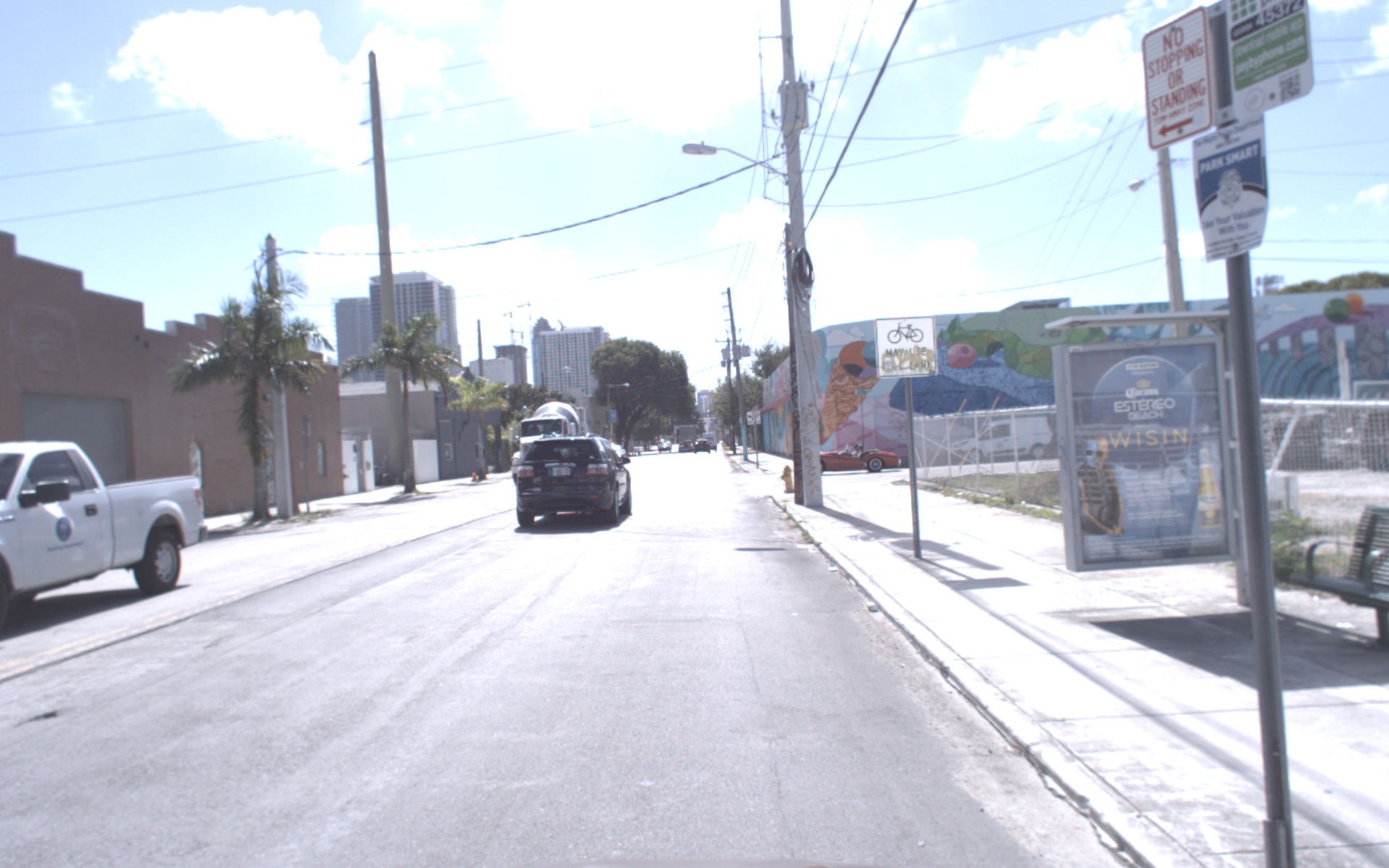}
    }
    \subfloat[]{%
        \includegraphics[width=.45\linewidth]{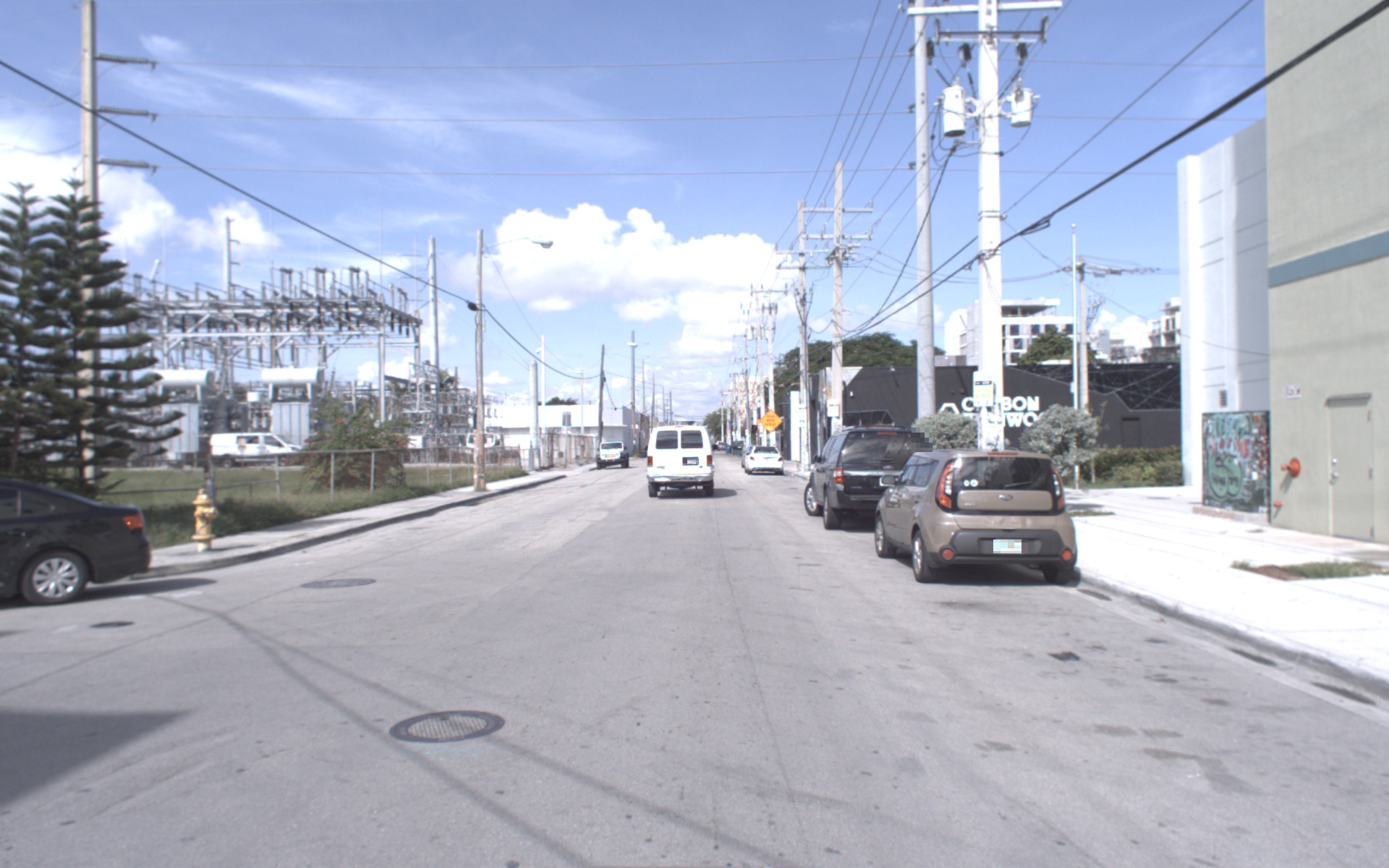}
    }
    \caption{\small Rapid Fluctuations in Performance Under Identical Weather Conditions: (a) Clip id: 00c561 shows a Streaming Average Precision (sAP) of 16.2\% using the StreamYOLO-s model, (b) Clip id: 395560 demonstrates a significantly higher sAP of 48.3\% under the same model and weather condition, illustrating the variability in model performance even under consistent environmental factors.}
    \label{fig:appendix_2}
\end{figure}

However, it's noteworthy that even within the same weather conditions, model performance varies significantly, with accuracy ranging from below 10\% to above 70\%.~Fig.~\ref{fig:appendix_2} illustrates this point by comparing frames from two video segments (Clip ids: {\footnotesize\texttt{00c561}} and {\footnotesize\texttt{395560}}) under identical weather conditions, where the performance difference of the same model on these segments is as high as 32.1\%.~This observation suggests the presence of other critical environmental factors that affect streaming perception, indicating that weather, while influential, is not the sole determinant of model performance.

\begin{table}[ht]
\centering
\footnotesize
\begin{tabular}{l|rrr}
\toprule
              & test & train & val\\[0pt]
\midrule
Sunny + Day   & 8    & 34    & 8\\[0pt]
Cloudy + Day  & 13   & 27    & 15\\[0pt]
Rainy + Day   & 1    & 1     & 0\\[0pt]
Rainy + Night & 1    & 0     & 0\\[0pt]
Sunny + Night & 1    & 3     & 1\\[0pt]
\bottomrule
\end{tabular}
\caption{\small Distribution of Weather Conditions in Testing, Training, and Validation Sets: This figure illustrates the frequency of different weather conditions in the testing, training, and validation sets of the Argoverse-HD dataset, providing an overview of the environmental variability within each dataset subset.}
\label{table:appendix_1}
\vspace{-1em}
\end{table}

\begin{table*}[ht]
\centering\footnotesize
\begin{tabular}{l|l|rrrrr|rrrr|rrrr}
\toprule
 &  & \multicolumn{5}{c}{StreamYOLO} & \multicolumn{4}{c}{LongShortNet} & \multicolumn{4}{c}{DAMO-StreamNet} \\[0pt] \hline
Clip ID & Weather & s 1x & m 1x & l 1x & l 2x & l still & s 1x & m 1x & l 1x & l high & s 1x & m 1x & l 1x & l high\\[0pt]
\midrule
1d6767 & Cloudy + Day & \bfred{20.9} & \bfred{22.8} & \bfred{24.9} & 7.0 & \bfred{26.7} & \bfred{20.9} & \bfred{23.4} & \bfred{25.0} & \bfred{36.4} & \bfred{21.3} & \bfred{24.6} & \bfred{26.0} & \bfred{34.2}\\[0pt]
5ab269 & Cloudy + Day & 25.6 & 30.0 & 31.6 & \bfred{6.9} & 33.3 & 25.2 & 29.5 & 31.4 & 40.1 & 26.9 & 29.0 & 31.7 & 41.2\\[0pt]
70d2ae & Cloudy + Day & 26.3 & 31.4 & 37.9 &  9.4 & 41.0 & 25.2 & 31.0 & 37.5 & 44.7 & 27.7 & 34.8 & 34.3 & 44.9\\[0pt]
337375 & Cloudy + Day & 24.8 & 24.8 & 33.4 & 17.1 & 35.3 & 27.2 & 27.9 & 34.7 & 38.0 & 26.4 & 37.5 & 28.8 & 39.1\\[0pt]
7d37fc & Cloudy + Day & 32.5 & 36.4 & 41.5 & 15.5 & 42.1 & 33.6 & 37.7 & 40.8 & 45.8 & 35.2 & 40.1 & 39.4 & 45.7\\[0pt]
f1008c & Cloudy + Day & 38.6 & 42.0 & 44.4 & 11.3 & 46.2 & 40.0 & 40.4 & 45.3 & 50.3 & 39.1 & 42.4 & 45.8 & 54.1\\[0pt]
f9fa39 & Cloudy + Day & 35.7 & 39.5 & 41.8 &  9.9 & 48.1 & 33.2 & 39.8 & 42.9 & 50.1 & 38.8 & 44.1 & 44.3 & 51.4\\[0pt]
cd6473 & Cloudy + Day & 40.0 & 45.7 & 44.0 & 11.3 & 52.7 & 36.6 & 47.3 & 47.3 & 54.0 & 40.2 & 44.6 & 47.9 & 54.7\\[0pt]
cb762b & Cloudy + Day & 36.4 & 41.3 & 44.3 & 10.8 & 44.8 & 36.9 & 41.4 & 44.4 & 57.7 & 40.9 & 44.8 & 43.7 & 57.6\\[0pt]
aeb73d & Cloudy + Day & 39.6 & 44.6 & 45.2 & 12.5 & 46.7 & 39.2 & 46.7 & 45.9 & 52.3 & 42.6 & 46.4 & 47.5 & 51.3\\[0pt]
cb0cba & Cloudy + Day & 48.3 & 47.5 & 52.1 & 13.8 & 50.9 & 46.0 & 47.5 & 50.4 & 55.5 & 47.1 & 47.7 & 51.5 & 59.4\\[0pt]
e9a962 & Cloudy + Day & 45.6 & 53.8 & 55.4 & 15.8 & 58.8 & 44.0 & 52.8 & 55.6 & 60.7 & 45.1 & 50.2 & 52.9 & 56.2\\[0pt]
2d12da & Cloudy + Day & 50.8 & 56.5 & 56.2 & 11.9 & 58.8 & 48.5 & 54.6 & 56.6 & 59.1 & 53.1 & 54.8 & 57.5 & 63.8\\[0pt]
85bc13 & Cloudy + Day & \bfgreen{56.2} & \bfgreen{56.8} & \bfgreen{60.1} & \bfgreen{19.5} & \bfgreen{62.1} & \bfgreen{55.3} & \bfgreen{58.2} & \bfgreen{59.2} & \bfgreen{63.5} & \bfgreen{54.9} & \bfgreen{58.3} & \bfgreen{59.6} & \bfgreen{67.3}\\[0pt]
\midrule
00c561 & Sunny + Day & \bfred{16.2} & \bfred{19.0} & \bfred{20.5} & \bfred{5.1} & \bfred{22.2} & \bfred{17.6} & \bfred{20.1} & \bfred{20.2} & \bfred{26.4} & \bfred{17.9} & \bfred{19.3} & \bfred{21.5} & \bfred{25.2}\\[0pt]
c9d6eb & Sunny + Day & 22.5 & 28.9 & 32.5 & 07.5 & 35.3 & 22.6 & 28.8 & 32.9 & 39.1 & 24.5 & 26.0 & 28.4 & 38.6\\[0pt]
cd5bb9 & Sunny + Day & 23.3 & 24.9 & 25.8 &  6.2 & 27.2 & 23.4 & 25.2 & 25.8 & 30.4 & 23.4 & 25.7 & 26.2 & 31.5\\[0pt]
6db21f & Sunny + Day & 24.1 & 26.4 & 27.0 &  6.7 & 28.9 & 23.3 & 27.0 & 27.0 & 34.7 & 25.1 & 28.0 & 28.7 & 37.0\\[0pt]
647240 & Sunny + Day & 27.1 & 29.3 & 31.2 & 07.8 & 34.1 & 26.5 & 30.1 & 31.5 & 38.8 & 26.9 & 32.0 & 32.0 & 38.4\\[0pt]
da734d & Sunny + Day & 30.2 & 33.4 & 37.0 &  8.8 & 39.9 & 29.2 & 34.4 & 37.5 & 42.6 & 34.2 & 35.7 & 38.2 & 43.1\\[0pt]
5f317f & Sunny + Day & 31.9 & 42.3 & 45.9 &  8.9 & 50.1 & 32.8 & 42.0 & 46.1 & 51.2 & 40.0 & 44.6 & 47.0 & 54.0\\[0pt]
395560 & Sunny + Day & 49.3 & 61.2 & 60.6 & 11.3 & \bfgreen{72.1} & 51.7 & 60.7 & 58.5 & 65.4 & 58.9 & 63.4 & 57.8 & 59.6\\[0pt]
b1ca08 & Sunny + Day & \bfgreen{60.0} & \bfgreen{62.1} & \bfgreen{68.4} & \bfgreen{22.4} & 67.9 & \bfgreen{61.7} & \bfgreen{61.4} & \bfgreen{67.7} & \bfgreen{70.6} & \bfgreen{59.6} & \bfgreen{65.0} & \bfgreen{67.7} & \bfgreen{68.6}\\[0pt]
\midrule
033669 & Sunny + Night & 18.0 & 23.5 & 25.7 &  6.6 & 27.4 & 18.5 & 23.6 & 25.1 & 27.6 & 21.8 & 22.7 & 23.8 & 27.5\\[0pt]
\midrule
Overall & -- & 29.8 & 33.7 & 36.9 & 34.6 & 39.4 & 29.8 & 34.1 & 37.1 & 42.7 & 31.8 & 35.5 & 37.8 & 43.3\\[0pt]
\bottomrule
\end{tabular}
\caption{\small Offline Evaluation Results on the Argoverse-HD Validation Dataset:~It records the sAP scores across the 0.50 to 0.95 range for each clip.~The optimal and worst results are highlighted in \bfgreen{green} and \bfred{red} font under the same weather conditions.~The notation ``l high" is used as an abbreviation for the resolution \(1200\times1920\), providing a concise representation of the data.}
\label{table:appendix_2}
\end{table*}

\begin{figure}[!t]
    \centering
    \subfloat[]{%
        \includegraphics[width=\linewidth]{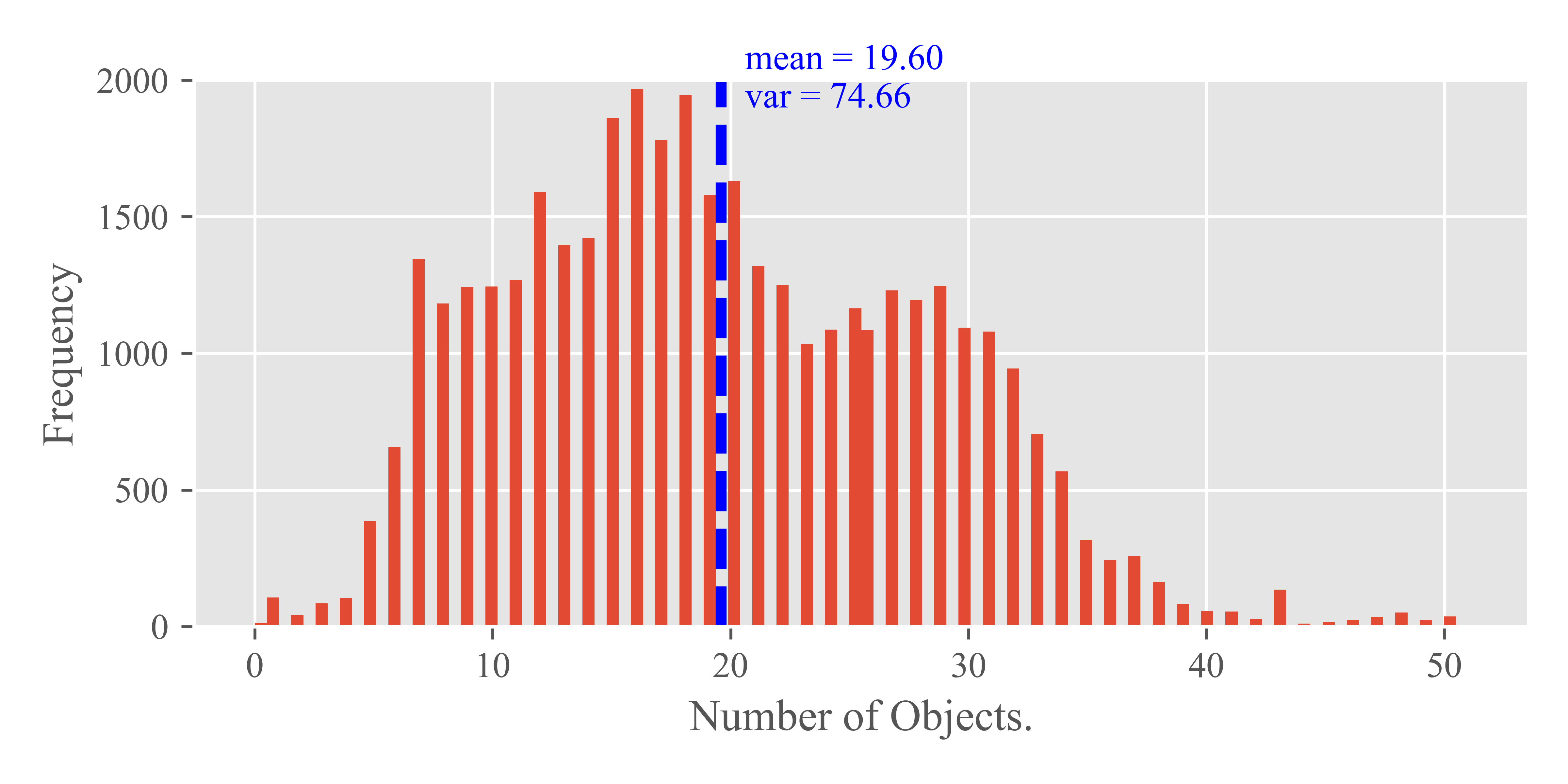}
    }\\
    \subfloat[]{%
        \includegraphics[width=\linewidth]{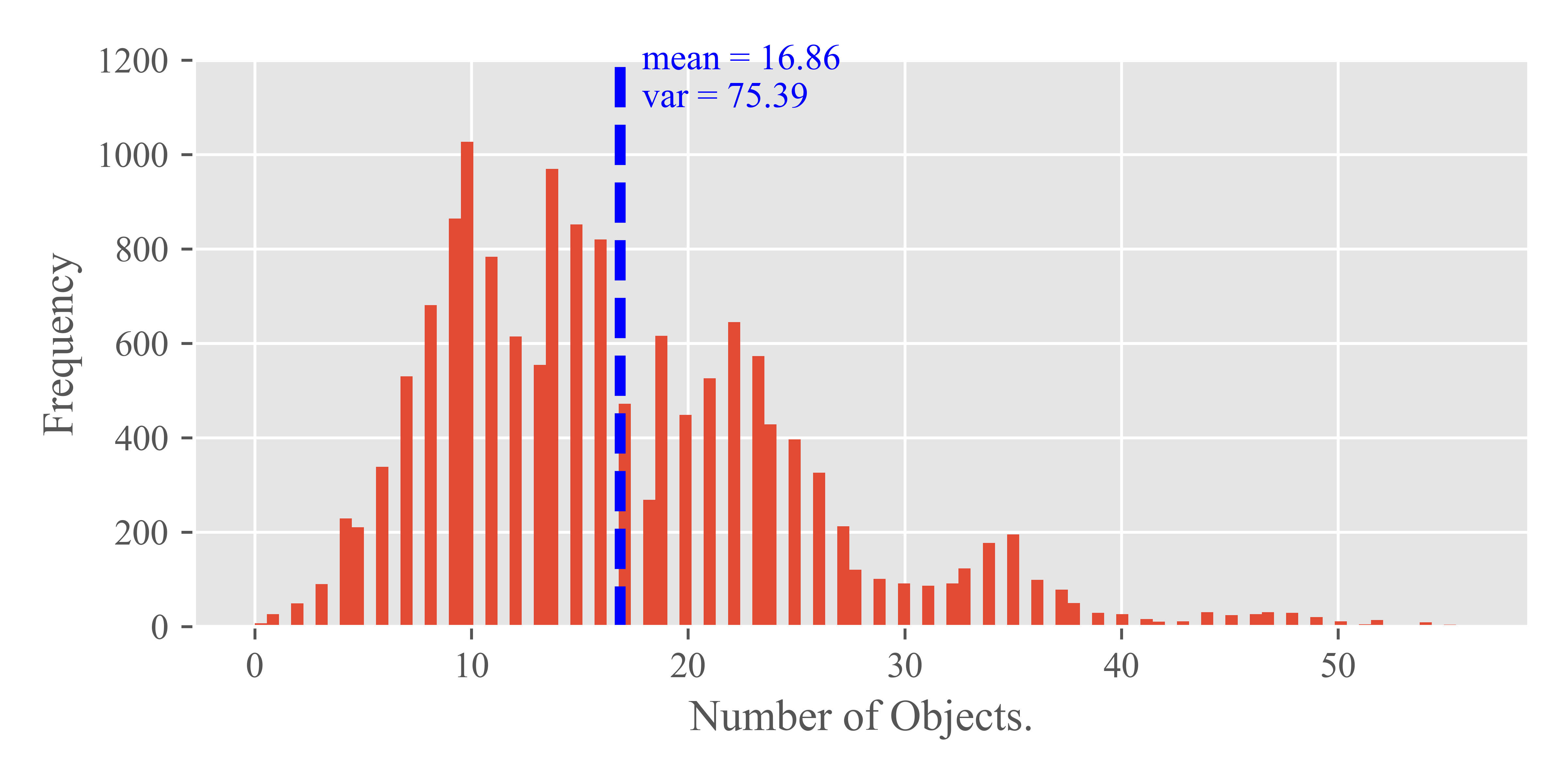}
    }
\caption{\small Histograms Depicting Object Quantity in the Argoverse-HD Dataset: This figure presents two histograms, (a) representing the distribution of the number of objects per frame in the training set of Argoverse-HD, and (b) showing the same distribution in the validation set.~These histograms provide a visual analysis of object frequency and variability within different sets of the dataset.}
    \label{fig:appendix_3}
\end{figure}

\subsection{Analysis of Object Quantity Impact}%
\label{sub:The quantity of objects}
To assess the impact of the number of objects on streaming perception, we conducted a statistical analysis of object counts per frame in the Argoverse-HD dataset, encompassing both training and validation sets.~The results of this analysis are depicted in Fig~\ref{fig:appendix_3}, which showcases a histogram representing the distribution of the number of objects in individual frames.~The variance in the distribution is notable, with values of \(74.66\) for the training set and \(75.39\) for the validation set, indicating significant fluctuation in the number of objects across frames.~Additionally, as shown in Tab.~\ref{table:appendix_2}, there is considerable variability in object counts across different video segments.~This observation led us to further investigate the potential correlation between object quantity and model performance fluctuations.

To explore this correlation, we calculated the average number of objects per frame for each segment within the Argoverse-HD validation set.~The findings, detailed in Tab.~\ref{table:appendix_3}, include the average object counts alongside Spearman correlation coefficients, which measure the relationship between object quantity and model performance.~The absolute values of these coefficients range from 1e-1 to 1e-2.~This range of correlation coefficients suggests that the number of objects present in the environment does not exhibit a strong or significant correlation with the performance of streaming perception models.~In other words, our analysis indicates that the sheer quantity of objects within the environment is not a predominant factor influencing the efficacy of streaming perception.

\begin{table}[ht]
\centering
\footnotesize
\begin{tabular}{l|crrr}
\toprule
Clip ID & Mean Obj $\uparrow$ & sYOLO & LSN & DAMO\\[0pt]
\midrule
1d6767 & 35.30 & 20.9 & 20.9 & 21.3\\[0pt]
7d37fc & 30.89 & 32.5 & 33.6 & 35.2\\[0pt]
da734d & 25.16 & 30.2 & 29.2 & 34.2\\[0pt]
cd6473 & 23.75 & 40.0 & 36.6 & 40.2\\[0pt]
5ab269 & 23.37 & 25.6 & 25.2 & 26.9\\[0pt]
cb762b & 23.31 & 36.4 & 36.9 & 40.9\\[0pt]
f1008c & 23.08 & 38.6 & 40.0 & 39.1\\[0pt]
e9a962 & 21.58 & 45.6 & 44.0 & 45.1\\[0pt]
70d2ae & 21.38 & 26.3 & 25.2 & 27.7\\[0pt]
2d12da & 19.33 & 50.8 & 48.5 & 53.1\\[0pt]
337375 & 18.19 & 24.8 & 27.2 & 26.4\\[0pt]
f9fa39 & 17.46 & 35.7 & 33.2 & 38.8\\[0pt]
aeb73d & 16.82 & 39.6 & 39.2 & 42.6\\[0pt]
6db21f & 16.30 & 24.1 & 23.3 & 25.1\\[0pt]
647240 & 14.18 & 27.1 & 26.5 & 26.9\\[0pt]
b1ca08 & 14.08 & 60.0 & 61.7 & 59.6\\[0pt]
85bc13 & 12.06 & 56.2 & 55.3 & 54.9\\[0pt]
033669 & 11.89 & 18.0 & 18.5 & 21.8\\[0pt]
00c561 & 10.06 & 16.2 & 17.6 & 17.9\\[0pt]
cb0cba & 10.04 & 48.3 & 46.0 & 47.1\\[0pt]
395560 & 10.00 & 49.3 & 51.7 & 58.9\\[0pt]
cd5bb9 & 8.95  & 23.3 & 23.4 & 23.4\\[0pt]
c9d6eb & 7.88  & 22.5 & 22.6 & 24.5\\[0pt]
5f317f & 6.92  & 31.9 & 32.8 & 40.0\\[0pt]
\midrule
Coefficient & -- & 0.052 & 0.035 & -0.020\\[0pt]
\bottomrule
\end{tabular}
\caption{\small Table~\ref{table:appendix_3} shows the analysis of the average number of objects per frame for each segment in the Argoverse-HD validation set, along with the Spearman correlation coefficients. These coefficients determine the relationship between the quantity of objects and the performance of streaming perception models. The coefficients range from 1e-1 to 1e-2, indicating a weak correlation. This data suggests that the total number of objects in the environment does not significantly affect the performance of streaming perception models, indicating that object quantity is not a primary factor that affects the efficacy of streaming perception tasks.}
\label{table:appendix_3}
\vspace{-4mm}
\end{table}

\subsection{Analysis of the Proportion of Small Objects}
\label{sub:The quantity of small objects}
The influence of small objects on perception models, particularly in autonomous driving scenarios, has been underscored in studies like \cite{mahaur2023small} and \cite{yang2022real}.~In such scenarios, even minor shifts in viewing angles can cause notable relative displacement of small objects, posing a challenge for perception models in processing streaming data effectively.~This observation prompted us to closely examine the proportion of small objects in the environment.

\begin{figure}[!t]
    \centering
    \subfloat[]{%
        \includegraphics[width=.9\linewidth]{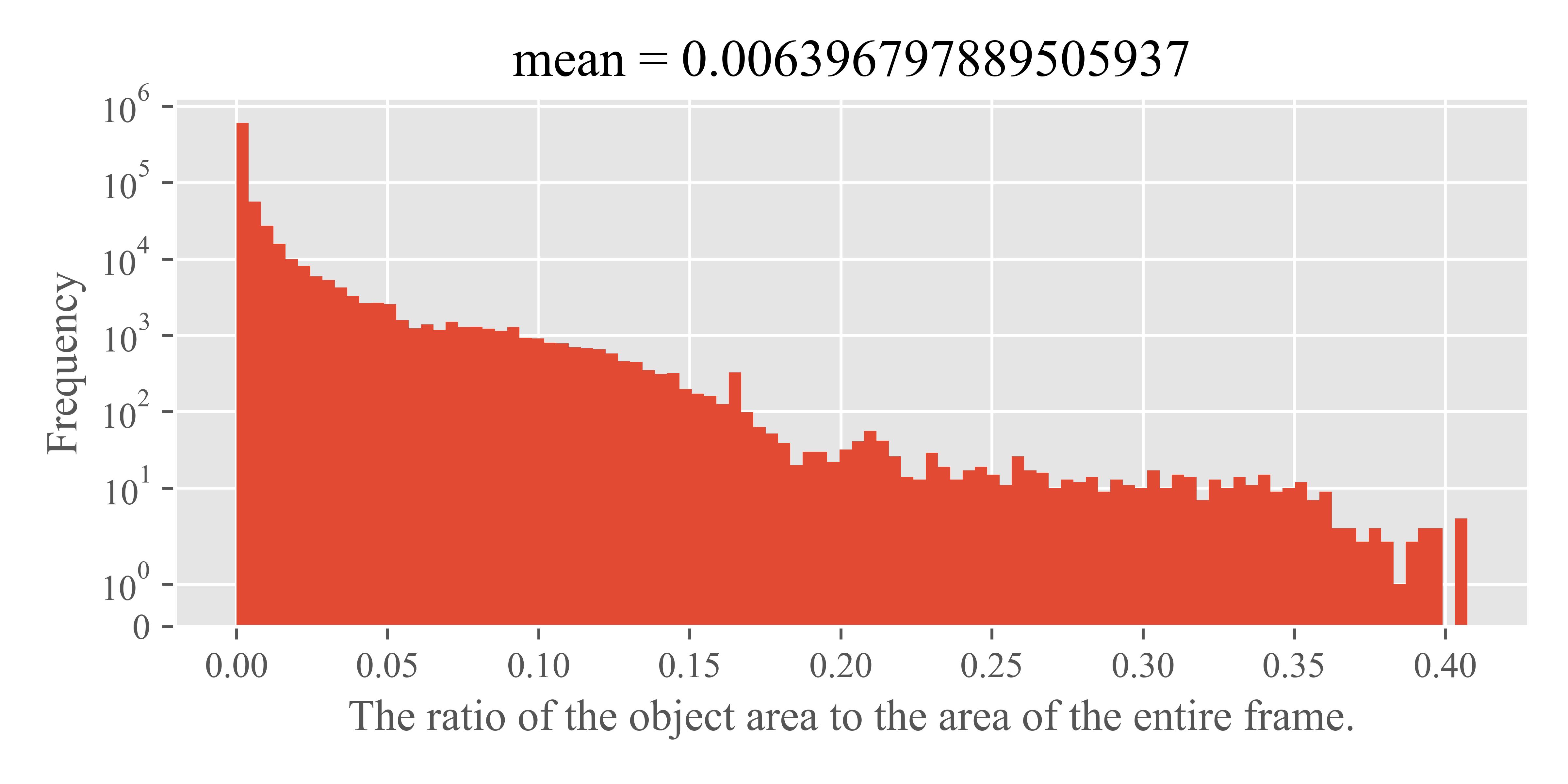}
    }\\
    \subfloat[]{%
        \includegraphics[width=.9\linewidth]{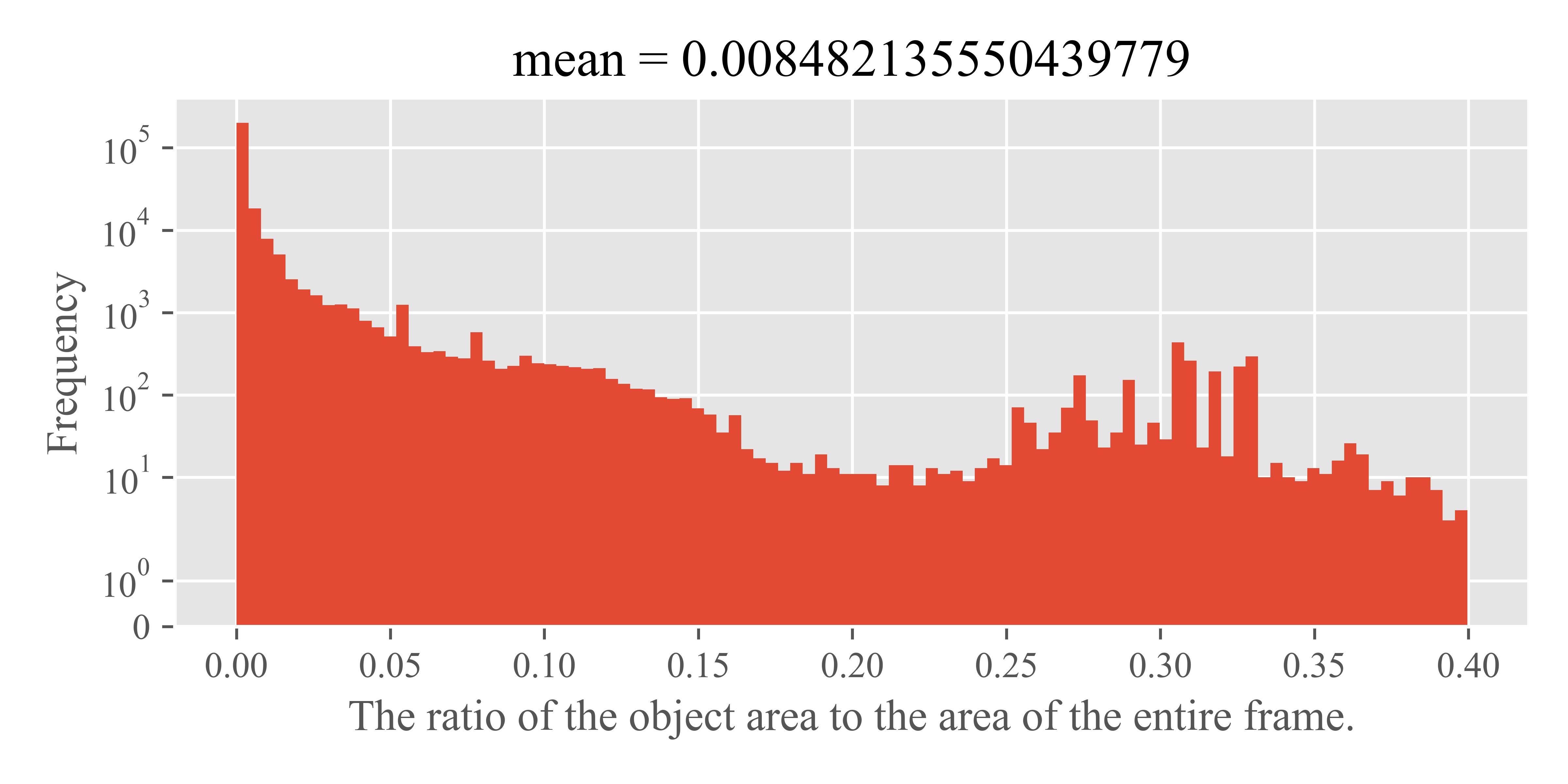}
    }
    \caption{\small Histograms of Object Area Proportions in Argoverse-HD Dataset: This figure showcases two histograms depicting the proportion of area occupied by objects relative to the entire frame, for (a) the training set and (b) the validation set of the Argoverse-HD dataset.~These histograms provide insights into the spatial distribution and size variation of objects within the frames of the dataset.}
    \label{fig:appendix_4}
\end{figure}

To begin, we analyzed the area ratios of objects in both the training and validation sets of the Argoverse-HD dataset.~This involved calculating the ratio of the pixel area covered by an object's bounding box to the total pixel area of the frame.~We visualized these ratios in histograms shown in Fig.~\ref{fig:appendix_4}.~The analysis revealed that the mean object area ratio is below 1e-2, indicating a substantial presence of small objects in the dataset.~For simplicity in subsequent discussions, we define objects with an area ratio less than 1\% as `small objects'.

Tab.~\ref{table:appendix_4} presents our findings on the proportion of small objects within the Argoverse-HD validation set.~Despite some variability in the overall number of objects and small objects, the proportion of small objects remains relatively stable, as reflected in the variance of their proportion.~This stability suggests that small objects are a consistent and prominent feature across various video segments, representing a persistent challenge of streaming perception.

\begin{table}[ht]
\centering
\footnotesize
\begin{tabular}{r|rrr}
\toprule
sid & \# obj $\uparrow$ & \# small obj & proportion\\[0pt]
\midrule
12 & 27829 & 24033 & 86\%\\[0pt]
3  & 16557 & 15937 & 96\%\\[0pt]
14 & 15058 & 14260 & 95\%\\[0pt]
15 & 12685 & 10229 & 81\%\\[0pt]
9  & 12618 & 11216 & 89\%\\[0pt]
5  & 12189 & 9509  & 78\%\\[0pt]
21 & 11801 & 10259 & 87\%\\[0pt]
18 & 11073 & 9856  & 89\%\\[0pt]
20 & 11068 & 10203 & 92\%\\[0pt]
7  & 10962 & 9707  & 89\%\\[0pt]
23 & 10961 & 9839  & 90\%\\[0pt]
2  & 10717 & 9700  & 91\%\\[0pt]
10 & 10706 & 9001  & 84\%\\[0pt]
22 & 10122 & 8846  & 87\%\\[0pt]
11 & 9965  & 8976  & 90\%\\[0pt]
4  & 9180  & 7989  & 87\%\\[0pt]
1  & 9068  & 8153  & 90\%\\[0pt]
24 & 8293  & 7830  & 94\%\\[0pt]
19 & 8068  & 6552  & 81\%\\[0pt]
17 & 4709  & 4230  & 90\%\\[0pt]
6  & 4420  & 3708  & 84\%\\[0pt]
16 & 7001  & 6508  & 93\%\\[0pt]
13 & 5654  & 5251  & 93\%\\[0pt]
8  & 3237  & 2449  & 76\%\\[0pt]
\midrule
mean & 10580 & 9343 & 87.96\%\\[0pt]
var & -- & -- & 0.0026\\[0pt]
\bottomrule
\end{tabular}
\caption{\small Distribution of Small Objects in the Argoverse-HD Validation Set: This figure illustrates the count of objects in each video segment of the Argoverse-HD validation set, specifically focusing on objects with an area proportion less than 1\%.~The chart provides a detailed view of the prevalence and distribution of smaller-sized objects across different video segments in the dataset.}
\label{table:appendix_4}
\end{table}

\subsection{Impact of Environmental Speed}%
\label{sub:The speed of environment}
In Sec.~\ref{sub:The quantity of small objects}, we highlighted how motion within the observer's viewpoint can affect the perception of small objects.~This observation leads us to consider that the speed of the environment could interact with the proportion of small objects.

To investigate the relationship between the environmental speed and the performance variability of streaming perception models, we categorized the validation dataset into three distinct environmental states: \emph{stop}, \emph{straight}, and \emph{turning}.~We then manually divided the dataset based on these states.~In this reorganized dataset, the clips with an ID's first digit as 0 exclusively represent the \emph{stop} state, while the digits 1 and 2 correspond to \emph{straight} and \emph{turning} states, respectively.

\begin{figure}[!t]
    \centering
    \subfloat[]{%
        \includegraphics[width=.82\linewidth]{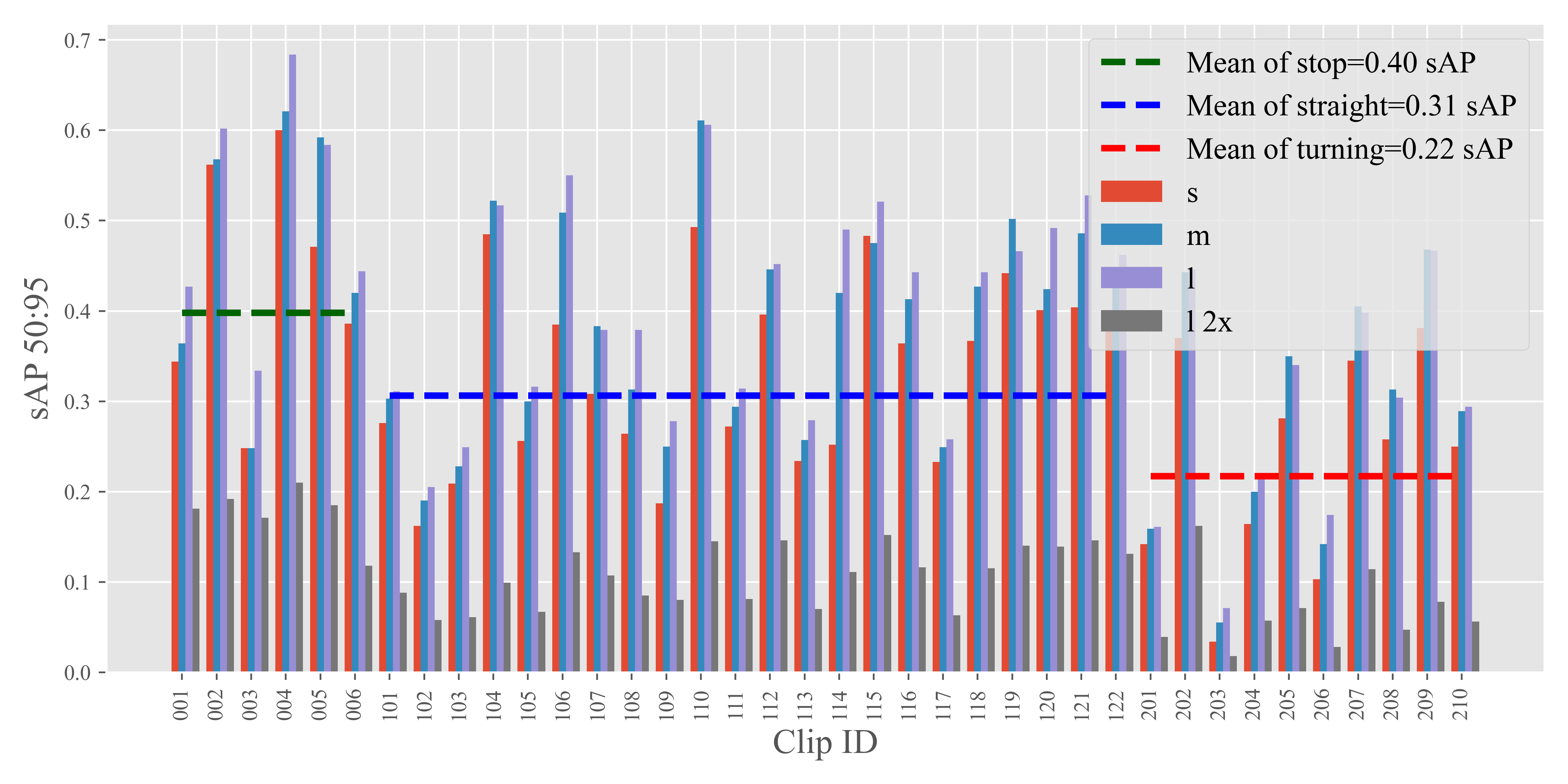}
    }\\
    \subfloat[]{%
        \includegraphics[width=.82\linewidth]{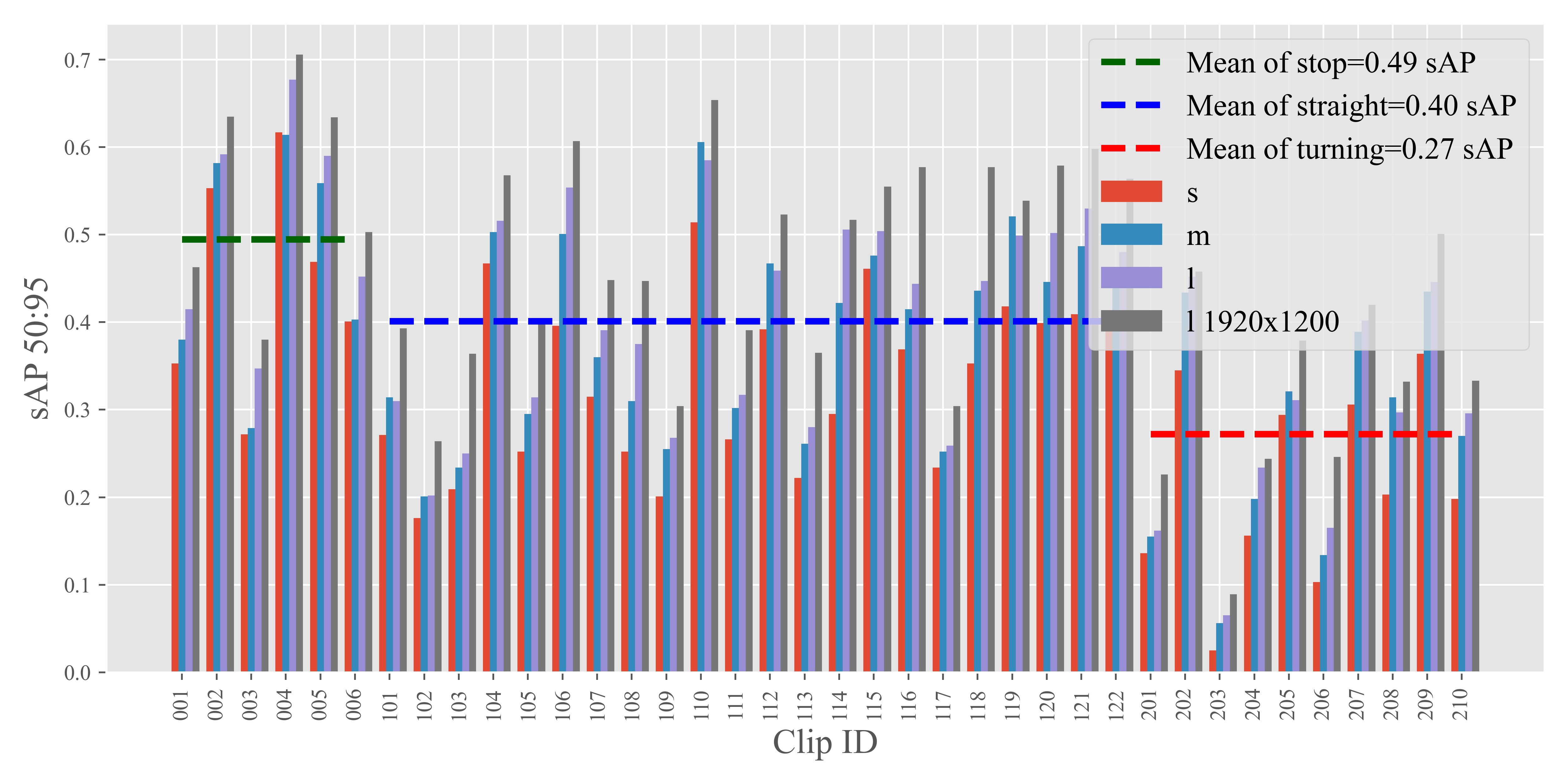}
    }\\
    \subfloat[]{%
        \includegraphics[width=.82\linewidth]{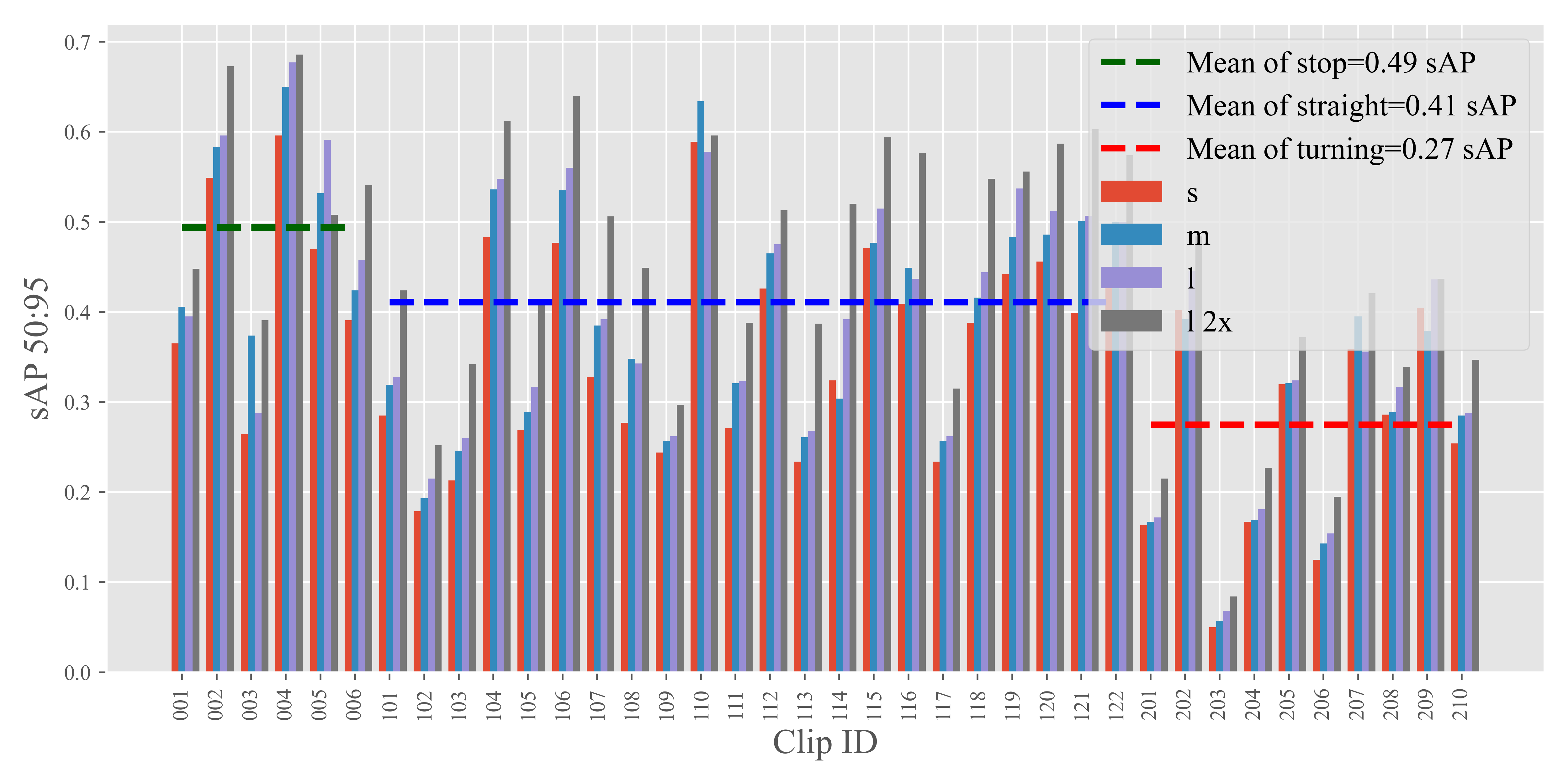}
    }\\   
\caption{\small Performance Analysis by Environmental Speed in Validation Segments: This figure displays the performance outcomes of three different models—(a) StreamYOLO, (b) LongShortNet, and (c) DAMO-StreamNet—across various segments of the Argoverse-HD validation set, categorized by environmental speed.~The charts provide a comparative view of how each model responds to different speeds in the environment, highlighting their effectiveness in varying dynamic conditions.}
    \label{fig:appendix_5}
    \vspace{-1em}
\end{figure}

Fig.~\ref{fig:appendix_5} showcases the performance of StreamYOLO, LongShortNet, and DAMO-StreamNet across each of these segments.~Additionally, the mean performance under each motion state is calculated and presented.~The data reveals a consistent pattern across all three models: the performance ranking in different environmental motion states follows the order of \emph{stop} being better than \emph{straight}, which in turn is better than \emph{turning}.~This trend indicates an association between the state of environmental motion and fluctuations.

Consequently, based on this analysis, we infer that the speed of the environment, particularly when considering the substantial proportion of small objects and their sensitivity to environmental dynamics, emerges as the most influential environmental factor in the context of streaming perception.

\section{More Experiment Results}%
\label{sec:More experiment results}

\subsection{Inference Time Analysis}
\label{sub:Inference Time}

This subsection supplements Section 4.4 of the main paper, where we previously discussed the performance of \methodname but did not extensively delve into its inference time characteristics.~To address this, Tab.~\ref{table:inference_time_supply} presents a detailed comparison of the inference times for each independent branch used in our model.~It is important to note that the inference times reported here may show variations when compared to those published by the original authors of the models.~This discrepancy is primarily due to differences in the hardware platforms used and the specific configurations of the corresponding models in our experiments.

An interesting observation from the results is that there are instances where \methodname exhibits a slower inference time compared to either the \emph{random} selection method or \emph{branch 1}.~This slowdown is attributed to the incorporation of the speed router in our sample routing mechanism.~Despite this, it is evident from the overall results that \methodname, employing the router strategy, still retains real-time processing capabilities across the various branches in the model bank.~Moreover, in certain scenarios, \methodname demonstrates even faster inference speeds than when using individual branches independently.~This detailed analysis underlines the dynamic and adaptive nature of \methodname in balancing between inference speed and accuracy, highlighting its capability to optimize streaming perception tasks in real-time scenarios.

\begin{table}[!t] \footnotesize
    \centering
    \setlength{\tabcolsep}{1.25mm}{
        \begin{tabular}{c|c c|c c}
            \toprule
            Branches & branch 0 & branch 1 & random & \methodname \\
            \midrule
            \(\text{DAMO}_\text{S + M}\)  & 29.26 & 33.65 & 36.61 & \bfgreen{33.22} \\
            \(\text{DAMO}_\text{S + L}\)  & 29.26 & 36.63 & \bfgreen{35.12} & 39.60 \\
            \(\text{DAMO}_\text{M + L}\)  & 33.65 & 36.63 & \bfgreen{37.30} & 37.61 \\
            \(\text{LSN}_\text{S + M}\)   & 22.08 & 25.88 & 24.79 & \bfgreen{21.47} \\
            \(\text{LSN}_\text{S + L}\)   & 22.08 & 31.24 & \bfgreen{21.49} & 30.48 \\
            \(\text{LSN}_\text{M + L}\)   & 25.88 & 31.24 & \bfgreen{24.75} & 29.05 \\
            \(\text{sYOLO}_\text{S + M}\) & 18.76 & 23.01 & 39.16 & \bfgreen{26.25} \\
            \(\text{sYOLO}_\text{S + L}\) & 18.76 & 27.85 & \bfgreen{24.04} & 29.35 \\
            \(\text{sYOLO}_\text{M + L}\) & 23.01 & 27.85 & 24.69 & \bfgreen{23.51} \\
            \bottomrule
    \end{tabular}}
    \caption{\small In-Depth Analysis of \methodname's Inference Time: This table presents a detailed comparison of inference times between the \emph{random} selection method and \methodname.~For ease of analysis, the optimal values in each comparison are highlighted in \bfgreen{green} font.~This highlighting assists in quickly identifying which method—\emph{random} or \methodname—achieves superior performance in terms of inference speed under various conditions.}
    \label{table:inference_time_supply}
\end{table}

\subsection{Statistic of model selection}
\label{sub:Statistic of model selection}

We also provide statistics on \methodname's selection of different models during both training and inference time in Tab.\ref{tab:statistic of model selection}. From the results, it can be observed that DAMO-StreamNet (M+L) exhibits a bias to select the second model during inference time, leading to a similar performance as DAMO-StreamNet L. However, under normal circumstances, \methodname can still dynamically choose the appropriate model based on input conditions.

\begin{table}
    \footnotesize
    \centering
    \begin{tabular}[c]{l|r|r|r|r} 
        \toprule
        Model & \multicolumn{2}{c|}{training time} & \multicolumn{2}{c}{inference time}\\
        Combination & Model 1 & Model 2 & Model 1 & Model 2\\
        \midrule
        SYOLO S+M & 14.24\% & 85.76\% &  5.95\% &  94.05\%\\
        SYOLO S+L & 10.98\% & 89.02\% &  4.83\% &  95.17\%\\
        SYOLO M+L & 37.53\% & 62.47\% & 94.67\% &   5.33\%\\
        \midrule
        LSN S+M   & 13.05\% & 86.95\% & 81.65\% &  18.35\%\\
        LSN S+L   &  7.28\% & 92.72\% & 17.26\% &  82.74\%\\
        LSN M+L   & 30.86\% & 69.14\% & 19.87\% &  80.13\%\\
        \midrule
        DAMO S+M  &  6.26\% & 93.74\% &  0.00\% & 100.00\%\\
        DAMO S+L  & 35.29\% & 64.71\% &  3.69\% &  96.31\%\\
        DAMO M+L  & 84.61\% & 15.39\% &  0.02\% &  99.98\%\\
        \bottomrule
    \end{tabular}
    \caption{The statistics of model selection by \methodname under different model choices during both training and inference time.}
    \label{tab:statistic of model selection}
\end{table}

\subsection{The comparison between Speed Router and \texorpdfstring{$\mathbb{E}[\Delta I_t]$}{TEXT}}
\label{sub:The comparison between Speed Router and mean of delta I_t}

\begin{table}[!t]\footnotesize
    \centering
    \setlength{\tabcolsep}{1.25mm}{
        \begin{tabular}{c|c|c}
            \toprule
            Model Bank & $\mathbb{E}[\Delta I_{t}]$ (sAP) & Speed Router (sAP) \\
            \midrule
            \(\text{StreamYOLO}_\text{S + M}\) & 31.5 & 32.6 (+1.1)\\
            \(\text{StreamYOLO}_\text{S + L}\) & 32.9 & 35.0 (+2.1)\\
            \(\text{StreamYOLO}_\text{M + L}\) & 34.2 & 34.6 (+0.4) \\
            \bottomrule
    \end{tabular}}
    \caption{\small Comparison of Speed Router and \(\mathbb{E}[\Delta I_t]\).~Where \(\mathbb{E}[\Delta I_{t}]\) means directly select model by the sign of \(\mathbb{E}[\Delta I_{t}]\) without using Speed Router.}
    \label{table:comparison_of_speed_router_and_E_I_t}
\end{table}

We also consider a special case id Tab.\ref{table:comparison_of_speed_router_and_E_I_t}, where the model selection only base on the mean of \(\Delta I_t\) without using Speed Router, which is denoted as \(\mathbb{E}[\Delta I_t]\).~To be specific, the larger model is selected when \(\mathbb{E}[\Delta I_t] > 0\) and minor model is selected otherwise. Unlike Tab.\red{5} in the main text, both methods here are trained for 5 epoch using LoRA fine-tuning.~From the results in Tab.\ref{table:comparison_of_speed_router_and_E_I_t}, it can be seen that our proposed Speed Router has significant advantages compared to directly using \(\mathbb{E}[\Delta I_t]\) to select branches.

\begin{table}
    \centering \footnotesize
    \begin{tabular}[c]{l|r|r|r}
        \toprule
        & \(=0\) & \(>0\) & \(<0\)\\
        \midrule
        train  & 0.24\% & 48.22\% & 51.55\% \\
        test   & 0.30\% & 49.85\% & 49.85\% \\
        val    & 0.17\% & 49.18\% & 50.66\% \\
        \bottomrule
    \end{tabular}
    \caption{Statistics of the sign of \(\mathbb{E}(\Delta I_t)\) over Argoverse-HD.}\label{tab:statistic of mean of Delta I_t}
\end{table}

Furthermore, in Tab.\ref{tab:statistic of mean of Delta I_t}, we also conducted the statistic the sign of \(\mathbb{E}[\Delta I_t]\) on the Argoverse-HD. Results with absolute values less than 1e-6 were considered equal to 0. The results reveal that evenly distributing training across models did not effectively adapt them to varying speeds as our Speed Router did.

\section{More Details of \methodname}
\label{sec:More implementation details}

\begin{table}[htbp] \footnotesize
    \centering
    \setlength{\tabcolsep}{1.25mm}{
        \begin{tabular}{l|c|r}
            \toprule
            Model                           & Scale & \# of params\\
            \midrule
            \multirow{3}{*}{StreamYOLO}     & S     &  9,137,319 \\
                                            & M     & 25,717,863 \\
                                            & L     & 54,914,343 \\
            \midrule
            \multirow{3}{*}{LongShortNet}   & S     &  9,282,103 \\
                                            & M     & 25,847,783 \\
                                            & L     & 55,376,515 \\
            \midrule
            \multirow{3}{*}{DAMO-StreamNet} & S     & 18,656,357 \\
                                            & M     & 50,129,333 \\
                                            & L     & 94,156,945 \\
            \bottomrule
    \end{tabular}}    
    \caption{\small Parameter Count of Selected Pre-trained Models: This table lists the number of parameters for each pre-trained model chosen for our analysis.~It provides a quantitative overview of the complexity and size of the models, facilitating a comparison of their computational requirements.}
    \label{table:params_count}
    \vspace{-1em}
\end{table}

\subsection{Pre-trained Model Selection}
\label{sub:Pre-trained model selection}

As outlined in the main paper, our implementation of \methodname incorporates three existing models as branches within the Model Bank \(\mathcal{P}\): StreamYOLO\cite{yang2022real}, LongShortNet\cite{li2023longshortnet}, and DAMO-StreamNet\cite{he2023damo}.~These models were selected due to their specialized features and proven effectiveness in streaming perception tasks.~StreamYOLO is unique for its two additional pre-trained weight variants, each tailored for different streaming processing speeds.~This feature allows for adaptable performance depending on the speed requirements of the streaming task.~In contrast, LongShortNet and DAMO-StreamNet are equipped with pre-trained weights optimized for high-resolution image processing, making them suitable for scenarios where image clarity is paramount.

To ensure a diverse and versatile range of options within the Model Bank, our implementation of \methodname selectively utilizes the Small (S), Medium (M), and Large (L) variants of the pre-trained weights from each model.~This choice enables a balanced mix of processing speeds and resolution handling capabilities, catering to a wide range of streaming perception scenarios.~The specific details regarding the number of parameters for these pre-trained models can be found in Tab.\ref{table:params_count}, which provides a comparative overview to help in understanding the computational complexity for different tasks.

\subsection{Setting of Hyperparameters}
\label{sub:hyperparameters}
For all our experiments, we maintained consistent training hyperparameters to ensure comparability and reproducibility of results.~The experiments were executed on four RTX 3090 GPUs.~Considering the need for selecting the optimal branch model for each sample during the routing process, we established a batch size of \(4\), effectively allocating one sample to each GPU for parallel computation.

In alignment with the configuration used in StreamYOLO, we employed Stochastic Gradient Descent (SGD) as our optimization technique.~The learning rate was set to \(0.001 \times \text{BatchSize} / 64\), adapting to the batch size proportionally.~Additionally, we incorporated a cosine annealing schedule for the learning rate, integrated with a warm-up phase lasting one epoch to stabilize the initial training process.

Regarding data preprocessing, we ensured uniformity by resizing all input frames to \(600 \times 960\) pixels.~This standardization was crucial for maintaining consistency across different datasets and ensuring that our model could generalize well across various input dimensions.

\section{Details of experiment on NuScenes-H dataset}
\label{sub:experiments_on_nuscenes_h}

To meet the requirements of streaming perception tasks, nuScenes-H~\cite{wang2023are} enhances the commonly used autonomous driving perception dataset nuScenes~\cite{caesar2020nuscenes} by increasing the annotation frequency from 2Hz to 12Hz. While nuScenes encompasses data from three modalities—Camera, LiDAR, and Radar—nuScenes-H provides dense 3D object annotations exclusively for the 6 sensors of Camera modality.

As mentioned in the main text, we trained and evaluated \methodname on the nuScenes-H dataset. To accommodate the requirements for 2D object detection, the 3D object annotations in nuScenes-H are converted to 2D using publicly available conversion scripts. All experiments were conducted exclusively using the \texttt{CAM\_FRONT} viewpoint. The dataset partition details are summarized in Tab.~\ref{tab:statistics_of_nuscenes_h}, which includes the number of video clips, video frames, and the instance counts for each object category within the subsets. As it shows in Tab.~\ref{tab:statistics_of_nuscenes_h}, limited or even absent annotation for some categories resulted in lower overall test performance. For clarity, this dataset is referred to as nuScenes-H 2D as follows.

Before training \methodname, YOLOX~\cite{ge2021yolox} was trained for 80 epochs on nuScenes-H 2D to obtain pretrained weights. These weights were then used to initialize the branch models within \methodname. During the training of \methodname, each individual branch was trained for 10 epochs, followed by 5 epochs of training for the router. All other training settings were consistent with those described in the main text. As indicated by the experimental results presented, \methodname maintains strong selection capabilities across different branches on other datasets, demonstrating its adaptability under practical application conditions.

\begin{table}
    \centering \footnotesize
    \begin{tabular}[c]{l|r|r}
        \toprule
                             & train set & test set\\
        \midrule
        \# of video clips    &       120 &       30\\
        \# of frames         &     26705 &     6697\\
        \midrule
        \# of anno           &    225346 &    71819\\
        adult                &     32200 &    13920\\
        child                &        22 &      142\\
        wheelchair           &         0 &        0\\
        stroller             &         0 &      174\\
        personal\_mobility   &         0 &        2\\
        police\_officer      &         0 &        0\\
        construction\_worker &      1573 &      362\\
        animal               &        22 &        0\\
        car                  &    100487 &    25356\\
        motorcycle           &      4958 &      330\\
        bicycle              &      1844 &     1248\\
        bus.bendy            &       531 &      283\\
        bus.rigid            &      4854 &     1161\\
        truck                &     21801 &     4934\\
        construction         &      2154 &     1200\\
        emergency.ambulance  &        61 &        0\\
        emergency.police     &       112 &        0\\
        trailer              &      6799 &      805\\
        barrier              &     33058 &    10568\\
        trafficcone          &      8654 &    10096\\
        pushable\_pullable   &      5191 &      649\\
        debris               &       666 &      348\\
        bicycle\_rack        &       359 &      241\\
        \bottomrule
    \end{tabular}
    \caption{Dataset partition of nuScenes-H 2D, includes the number of video clips (\# of video clip), video frames (\# of video clip), and the instance counts for each object category (\# of anno) within the subsets.}
    \label{tab:statistics_of_nuscenes_h}
\end{table}

\end{document}